\documentclass{article}

\PassOptionsToPackage{numbers,sort&compress}{natbib}
\usepackage[main, final]{neurips_2026}

\usepackage[utf8]{inputenc} 
\usepackage[T1]{fontenc}    
\usepackage{hyperref}       
\usepackage{url}            
\usepackage{booktabs}       
\usepackage{amsfonts}       
\usepackage{nicefrac}       
\usepackage{microtype}      
\usepackage{xcolor}         

\usepackage{graphicx}
\usepackage{amsmath}
\usepackage{multirow}
\usepackage{algorithm}
\usepackage{algorithmic}
\usepackage{subcaption}
\usepackage{array}
\usepackage{colortbl}
\usepackage{wrapfig}

\newcommand{\red}[1]{\textcolor{red}{#1}}
\newcommand{\blue}[1]{\textcolor{blue}{#1}}
\newcommand{\vs}{\textit{vs.}\ }

\title{Controllable Multi-label Video Safety Detection via Adaptive Tversky Policy Optimization}

\author{%
  \textbf{Guangyu Yang}$^{1,2,*,\dagger}$ \qquad
  \textbf{Jingbiao Mei}$^{1,\dagger}$ \qquad
  \textbf{Mingsheng Sun}$^{3,\dagger}$ \qquad
  \textbf{Jinghong Chen}$^{1}$ \\[0.4em]
  \textbf{Yingtong Bu}$^{2}$ \qquad
  \textbf{Pengda Qin}$^{4}$ \qquad
  \textbf{Da Chen}$^{5,\ddagger}$ \qquad
  \textbf{Bill Byrne}$^{1,\ddagger}$ \\[0.6em]
  $^{1}$University of Cambridge \qquad
  $^{2}$Xiaohongshu Inc. \qquad
  $^{3}$AntGroup \\[0.3em]
  $^{4}$Tencent Company, China \qquad
  $^{5}$University of Bath \\[0.6em]
  \texttt{gy266@cam.ac.uk} \quad
  \texttt{jm2245@cam.ac.uk} \quad
  \texttt{sunms5513@gmail.com} \\[0.2em]
  \texttt{jc2124@cam.ac.uk} \quad
  \texttt{buyingtong819@gmail.com} \quad
  \texttt{derekpdqin@tencent.com} \\[0.2em]
  \texttt{da.chen@bath.edu} \qquad
  \texttt{wjb31@cam.ac.uk}
}

\begin{document}

\maketitle
\begingroup
\renewcommand{\thefootnote}{\fnsymbol{footnote}}
\footnotetext[1]{Work done during internship at Xiaohongshu Inc.}
\footnotetext[2]{Equal contribution.}
\footnotetext[3]{Corresponding authors.}
\endgroup


\begin{abstract}
  The rapid growth of video-based social media has increased users’ exposure to harmful content, creating a need for reliable automated video safety detection. Although recent Vision-Language Models (VLMs) show strong video understanding capabilities, existing harmful video detection systems face two key limitations: they typically reduce safety detection to binary classification, overlooking the inherently multi-label nature of unsafe videos, and they rely on static training objectives that do not support controllable precision-recall trade-offs, though the desired operating point may vary across moderation pipelines and unsafe categories. To address these gaps, we propose Adaptive Tversky Policy Optimization (ATPO), a reinforcement learning framework for Multi-label Video Safety Detection (Multi-VSD). ATPO introduces the Adaptive Tversky Reward (ATR), which dynamically adjusts false-positive and false-negative penalties during training to enable controllable precision–recall trade-offs. Experiments on SafeWatch-Bench and XD-Violence show that ATPO substantially improves multi-label performance, increasing the Jaccard Index from 40.66 to 75.44 on SafeWatch-Bench-Real. Moreover, ATR enables reliable steering of the precision–recall operating point, supporting deployment scenarios with heterogeneous policy requirements. Code and checkpoints are provided at \url{https://bruceyg.github.io/ATPO-project-page/}.
\end{abstract}

\section{Introduction}
\label{sec:intro}


The rapid growth of video-based social media platforms has led to an increasing amount of harmful video content\footnote{\red{This paper contains content for demonstration purposes that may be disturbing for some readers.}}.
Recent advances in Vision-Language Models (VLMs) have enabled strong performance on video understanding tasks~\cite{maaz-etal-2024-video,lin-etal-2024-video,bai2025qwen25vltechnicalreport,bai2025qwen3vltechnicalreport,feng2025videor,yan2025videochatr}, and there is a growing amount of work using VLMs for video safety detection~\cite{zhang2024holmes,atang2024hawk,Ye_2025_CVPR,huang2025vadr,ICLR2025_beac6bfb}. However, existing work on harmful video detection typically formulates the task as binary classification, distinguishing safe from unsafe content while treating all unsafe categories as a single class. Such formulations overlook the multi-class structure of unsafe content and are limited in their ability to provide the fine-grained and interpretable classification outcomes required by human content moderators~\cite{Calabrese2024StrucuturedExplanationsHumanMod,EXPOHM2026Mei}. In practice, \textbf{video safety detection is inherently multi-label}: unsafe videos often involve overlapping categories (e.g., violence and abuse) due to their rich and complex semantics.


Moreover, prior approaches typically optimize a static training objective and evaluate models at fixed operating settings~\cite{11094660,pmlr-v267-huang25ad,huang2025vadr}. This setup does not explicitly enable \textbf{controllable precision–recall trade-offs}, and is therefore misaligned with real-world deployment settings. In practice,  safety detection systems are often integrated as guardrails within multi-stage moderation pipelines, where the desired precision-recall operating point depends critically on policy requirements. For example, guardrail systems designed for children must prioritize high recall to minimize harmful exposure. In contrast, general-purpose moderation systems require a more balanced trade-off between precision and recall~\cite{10.5555/3716662.3716795,gongane2022detection}. Importantly, the optimal precision-recall trade-off may also differ across unsafe categories. Multi-label safety detection inherently involves heterogeneous risk profiles. Certain categories (e.g., misinformation) may prioritize higher precision to reduce over-censorship, while others (e.g., violent or extremist content) may prioritize higher recall to minimize the risk of harmful exposure.

\begin{figure}[tb]
  \centering
  \includegraphics[height=5.5cm]{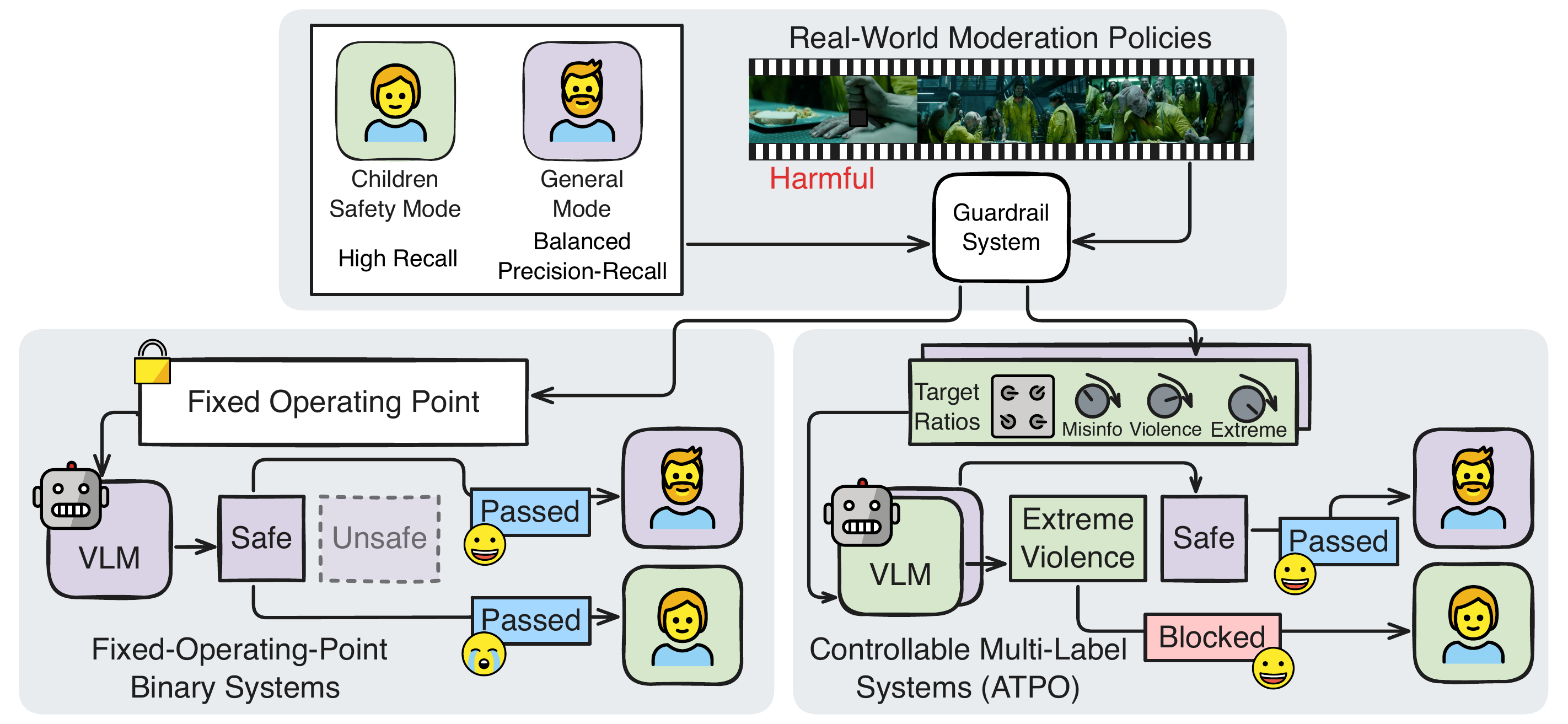}
  \caption{Real-world moderation policies require different precision–recall trade-offs: fixed-operating-point binary systems may miss harmful content and incorrectly pass it to sensitive audiences, whereas our proposed ATPO enables category-aware precision–recall control in multi-label video safety detection.}
  \label{fig:teaser}
\end{figure}


To address these gaps, we propose Adaptive Tversky Policy Optimization (ATPO), a Reinforcement Learning (RL) framework built upon Group Relative Policy Optimization (GRPO)~\cite{deepseek-math}. We first move beyond binary supervision by designing fine-grained multi-label safety training and rewards for GRPO, and systematically study reward formulations for multi-label optimization. Building on this, we draw inspiration from feedback loops in control theory to develop the Adaptive Tversky Reward (ATR). ATR reformulates the Tversky coefficients~\cite{https://doi.org/10.1037/0033-295X.84.4.327}, which govern False Positive (FP) and False Negative (FN) penalties, as feedback-driven variables that dynamically adjust during training based on observed model errors. It enables controllable and category-aware safety behavior within a unified multi-label detection model. Experiments on SafeWatch-Bench~\cite{ICLR2025_beac6bfb} and XD-Violence~\cite{Wu2020not} show that ATPO substantially improves Multi-label Video Safety Detection (Multi-VSD) over large general-purpose VLMs while maintaining strong binary performance. Furthermore, ATPO provides reliable control over the precision–recall trade-off both globally and at the category level, offering a practical mechanism for deployment.

Our contributions can be summarized as:
\begin{enumerate}
    \item We formulate multi-label harmful video detection under RL-based fine-tuning and systematically study reward formulations for optimizing multi-label safety objectives.
    \item We propose Adaptive Tversky Policy Optimization (ATPO), a feedback-driven RL framework that dynamically adjusts false-positive and false-negative penalties during training via the Adaptive Tversky Reward (ATR). Extensive experiments show that ATPO substantially improves multi-label video safety detection, increasing the Jaccard Index from 40.66 to 75.44 (Section~\ref{subsec:exp_main} and \ref{subsec:reward_comp}).

    \item Our experiments show that ATR enables reliable control over the precision–recall trade-off both globally and at the category level, supporting deployment scenarios with different risk sensitivities (Section~\ref{subsec:gcontrol_exp}). 
\end{enumerate}

\section{Related Work}
\label{sec:related_work}

Prior research on video safety detection has largely been studied through the paradigm of Video Anomaly Detection (VAD)~\cite{Sultani_2018_CVPR,liu2018ano_pred,Wu2020not,luo2017revisit,luo2019video,park2020learning}, where the objective is to determine whether a video, typically recorded by surveillance cameras, contains abnormal events. Early approaches commonly formulated VAD as a binary classification problem that distinguishes normal from anomalous activities~\cite{Sultani_2018_CVPR}. More recently, a line of work has leveraged Vision-Language Models (VLMs) for video anomaly detection and understanding~\cite{Ye_2025_CVPR,zhang2024holmes,11094660,Zanella2024HarnessingLL,huang2025vadr}. For example, Holmes-VAD~\cite{zhang2024holmes} trains a VLM with temporal sampling to detect anomalies and generate explanations for abnormal events in long videos, while VAD-R1~\cite{huang2025vadr} introduces Video Anomaly Reasoning (VAR), a structured chain-of-thought pipeline further optimized with an RL-based AVA-GRPO algorithm. Beyond binary detection, recent work has begun to study fine-grained video safety and anomaly recognition. Ex-VAD~\cite{pmlr-v267-huang25ad} extends VAD to recognize multiple abnormal event categories using vision-language supervision, but formulates the task as multi-class classification, where each video is assigned a single abnormal category. Open-Vocabulary VAD~\cite{wu2024openvocabularyvideoanomalydetection,11094336} uses pretrained VLMs and textual semantics to detect both seen and unseen anomaly categories. More recently, SafeWatch-Bench~\cite{ICLR2025_beac6bfb} explicitly frames video safety detection as a multi-category moderation task. Despite this progress, most prior methods either focus on binary anomaly detection or assign a single abnormal category to each video. In contrast, we study Multi-label Video Safety Detection (Multi-VSD), where multiple harmful categories may co-occur in the same video.


Reinforcement learning has also been explored for improving the reasoning and perception capabilities of VLMs for videos~\cite{feng2025videor,yan2025videochatr,li2025videochatr1enhancingspatiotemporalperception,huang2025vadr,li2025reinforcement,li2025tempsampr1effectivetemporalsampling}. Recent works apply GRPO-style RL fine-tuning to video understanding tasks, where reward functions are designed to encourage stronger temporal reasoning, spatio-temporal perception, and instruction-following abilities. Video-R1~\cite{feng2025videor} introduces reinforcement fine-tuning for video reasoning and designs temporal rewards to improve the model's ability to reason over video frames. VideoChat-R1~\cite{yan2025videochatr,li2025videochatr1enhancingspatiotemporalperception} further studies reinforcement fine-tuning for video dialogue and spatio-temporal perception, using task-specific rewards for video question answering and temporal understanding. Other recent works investigate reward design and data efficiency for VideoLLM RL~\cite{li2025reinforcement}, and use RL to improve temporal sampling and temporal grounding in video LLMs~\cite{li2025tempsampr1effectivetemporalsampling}. For safety-related tasks, GuardReasoner-VL~\cite{liu2025guardreasonervl} trains VLM guardrails with reasoning SFT and GRPO, while VAD-R1~\cite{huang2025vadr} applies RL to structured video anomaly reasoning. These studies show that reinforcement learning provides an effective post-training mechanism for adapting VLMs to video reasoning and safety tasks. Building on these developments, we study RL-based reward design for Multi-VSD, introducing adaptive rewards that account for partial label correctness and asymmetric FP/FN costs to enable controllable precision-recall trade-offs during training.

Recent work also studies video safety beyond anomaly detection. For generative models, T2VSafetyBench~\cite{3737916.3739955} evaluates safety risks in text-to-video generation, while SAFREE~\cite{yoon2025safree} provides a training-free safeguard for text-to-image and video generation. For video and multimodal LLMs, VideoSafety-R1~\cite{sun2026evaluationdefenseadvancingsafety} combines alarm-token-guided SFT with safety-guided GRPO to improve defensive reasoning. SafeVid~\cite{wang2026safevid} constructs video-specific safety preference data and applies Direct Preference Optimization (DPO)~\cite{3666122.3668460}, whereas SEA~\cite{lu-etal-2025-sea} uses synthetic modality embeddings to enable low-resource safety alignment from textual data. Complementary security and privacy studies examine video-modality jailbreaks through VideoJail~\cite{hu2025videojail} and protective adversarial watermarks that disrupt unauthorized video annotation~\cite{11092973}. These directions address safe generation, safe-response alignment, or protection against misuse of video inputs. ATPO instead studies moderation of existing videos through fine-grained multi-label predictions. 

\section{Proposed Method}
\label{sec:method}

Our training pipeline consists of Supervised Fine-Tuning (SFT) warmup  followed by reinforcement learning (RL). We first introduce the task formulation and notation for Multi-label Video Safety Detection (Multi-VSD) (Section~\ref{subsec:prelim}). We study different reward formulations, including standard overlap-based rewards (Section~\ref{subsec:grpo}) and our proposed Adaptive Tversky Policy Optimization (ATPO) (Section~\ref{subsec:atr}). The overall pipeline is illustrated in Figure~\ref{fig:method}.

\subsection{Multi-label Video Safety Detection Formulation}
\label{subsec:prelim}
We formulate the task of Multi-label Video Safety Detection (Multi-VSD) by extending previous binary detection. Given a video $v$, the model predicts whether the content belongs to any of a predefined set of unsafe categories $\mathcal{C}=\{c_1,\dots,c_K\}$, where each $c$ represents a type of harmful content (e.g., violence, misinformation, extremism).
For each video, the model produces a predicted label set $P \subseteq \mathcal{C}$, while the ground-truth annotation is a label set $G \subseteq \mathcal{C}$. We denote True Positive (TP), False Positive (FP), and False Negative (FN) as
\begin{equation}
\begin{aligned}
TP = |P \cap G|, \qquad
FP = |P \setminus G|, \qquad
FN = |G \setminus P|.
\end{aligned}
\end{equation}
These quantities characterize the quality of the predicted label set and form the basis of the reward functions used for reinforcement learning and evaluation metrics.

\subsection{GRPO with Static Rewards}
\label{subsec:grpo}
\begin{figure}[tb]
  \centering
  \includegraphics[height=7.0cm]{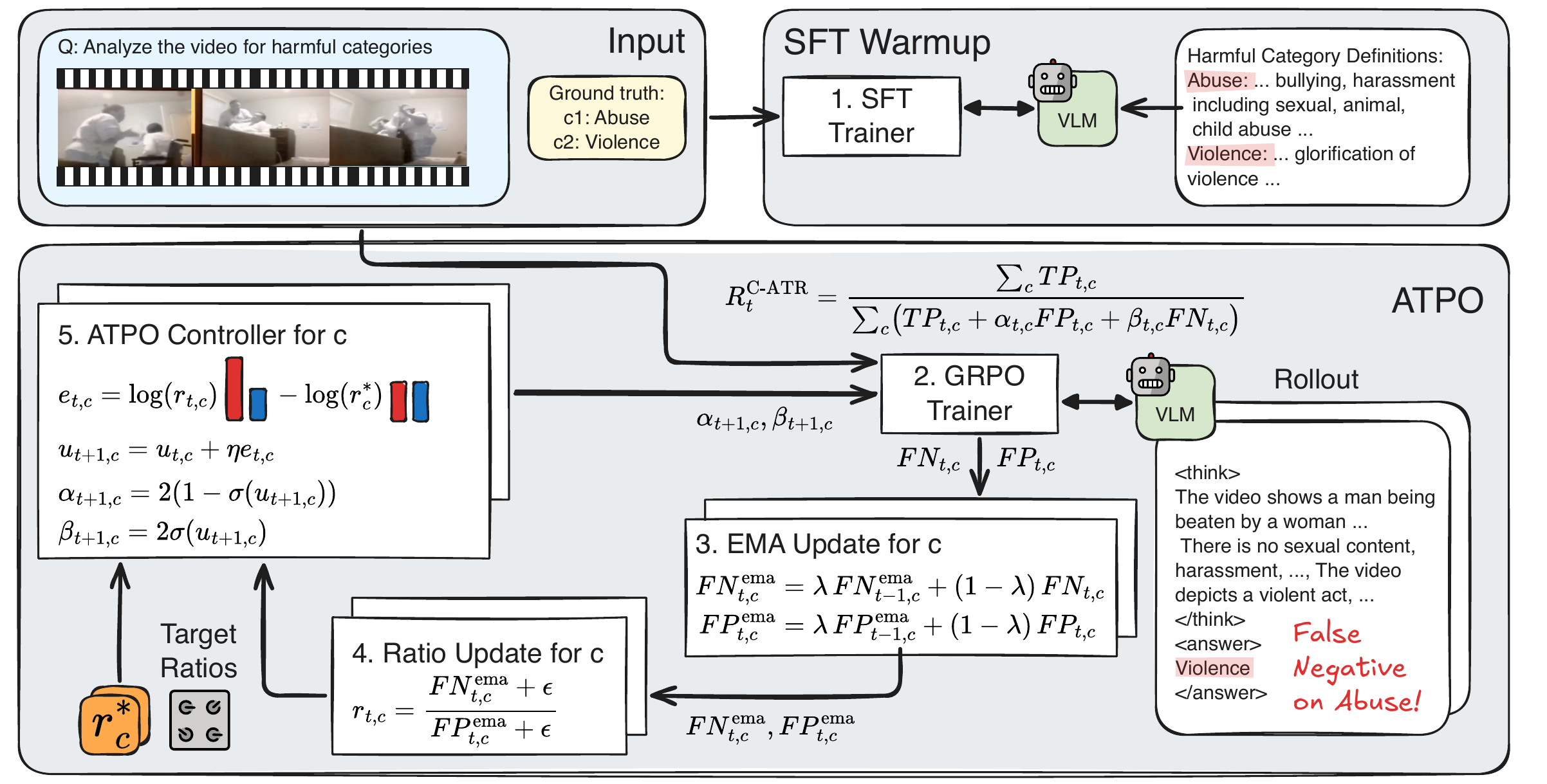}
  \caption{Overview of Adaptive Tversky Policy Optimization (ATPO). The training pipeline consists of five stages. (1) SFT warmup: The VLM is first supervised fine-tuned using harmful video category definitions. (2) GRPO training: The model generates predictions through GRPO rollouts, which are evaluated using Adaptive Tversky Reward (ATR) and model parameters are updated. (3) Category-wise \red{FNs} and \blue{FPs} from the rollouts are aggregated and tracked using EMA. (4) The \red{FN}-to-\blue{FP} ratio is updated using the EMA statistics. (5) ATPO controller compares the ratio with a user-specified target ratio $r^*_c$ and updates the Tversky coefficients $\alpha_{t+1, c}, \beta_{t+1, c}$ for the next training step.}
  \label{fig:method}
\end{figure}

Our training follows a two-stage pipeline consisting of SFT warmup followed by RL training, following recent works on post-training VLMs~\cite{feng2025videor,li2025reinforcement,zhang-etal-2025-improve}. We first perform SFT on the backbone VLM using a multi-label classification objective, as shown in Figure~\ref{fig:method}, where the target outputs correspond to the ground-truth label sets, without auxiliary annotations such as Chain-of-Thought (CoT)~\cite{10.5555/3600270.3602070} reasoning.

Following the SFT warmup stage, we further improve the multi-label prediction capability of VLMs using RL training. Group Relative Policy Optimization (GRPO)~\cite{deepseek-math} is adopted to optimize the model using reward functions defined over predicted and ground-truth label sets.
Given an input video $v$ and detection instruction $x$, the old policy model  $\pi_{\theta_{\text{old}}}$ generates $N$ sampled outputs $\{y_j\}_{j=1}^N$. Each output contains a predicted label set $P_j$. We evaluate each prediction using a reward function $R_j = \mathcal{R}(P_j, G)$ that measures its agreement with the ground-truth label set $G$.
Following GRPO, advantages are computed by normalizing rewards within the sampled group:
\begin{equation}
A_j = \frac{R_j - \text{mean}(\{R1,...,R_N\})}{\text{std}(\{R1,...,R_N\})}.
\end{equation}
The policy is then updated by the following GRPO objective, where $\mathbf{x} = (x, v)$:
\begin{equation}
\mathcal{L}_{\text{G}}(\theta)
=
-\frac{1}{N}\sum_{j=1}^N 
\Big[
\min\Big(
\frac{\pi_{\theta}(y_j \mid \mathbf{x})}{\pi_{\theta_{\text{old}}}(y_j \mid \mathbf{x})} A_j,
\text{clip}\Big(
\frac{\pi_{\theta}(y_j \mid \mathbf{x})}{\pi_{\theta_{\text{old}}}(y_j \mid \mathbf{x})},
1 - \epsilon, 1 + \epsilon
\Big) A_j
\Big)
- \beta' D_{\text{KL}}\!\left(\pi_{\theta} \,\|\, \pi_{\text{ref}}\right)
\Big]
\label{eq:grpo}
\end{equation}

\noindent\textbf{Exact Match reward.}
As a basic reward function, we consider Exact Match (EM), which assigns a reward of 1 only when the predicted label set exactly matches the ground-truth set:
\begin{equation}
\mathcal{R}^{\text{EM}}(P,G)
=
\mathbf{1}[P = G].
\end{equation}
While simple, EM is a strict objective that provides sparse feedback. As we show in Section~\ref{subsec:reward_comp}, such sparse rewards lead to suboptimal learning for Multi-VSD, motivating the use of partial reward functions based on TP, FP, and FN statistics.

\noindent\textbf{Partial rewards based on set overlap.}
The Jaccard Index (JI)~\cite{https://doi.org/10.1111/j.1469-8137.1912.tb05611.x} is defined as the Intersection-over-Union between $P$ and $G$:
\begin{equation}
\mathcal{R}^{\text{Jaccard}}(P,G)
= \frac{|P \cap G|}{|P \cup G|}
= \frac{TP}{TP + FP + FN}
\end{equation}
It treats false positives and false negatives symmetrically and corresponds to a special case of the Tversky Index with equal weights.

The Tversky Index~(TI)~\cite{https://doi.org/10.1037/0033-295X.84.4.327} generalizes the Jaccard Index by introducing asymmetric penalties:
\begin{equation}
\mathcal{R}^{\text{STR}}_{\alpha,\beta}(P,G)
=
\frac{TP}{TP + \alpha \, FP + \beta \, FN},
\label{eq:tversky}
\end{equation}
where $\alpha,\beta > 0$ control the penalties of precision and recall. We note that the coefficient $\beta$ here is unrelated to the KL regularization coefficient $\beta'$ in Equation~\ref{eq:grpo}. We refer to this reward function as Static Tversky Reward (STR) because the coefficients are fixed throughout training.
When $\alpha = \beta = 1$, STR reduces to the Jaccard reward. 
Increasing $\alpha$ emphasizes precision (penalizing FP more), while increasing $\beta$ emphasizes recall (penalizing FN more).

\subsection{Adaptive Tversky Policy Optimization (ATPO)}
\label{subsec:atr}

In practical detection pipelines, different deployment scenarios may require different precision-recall trade-offs. Instead of fixing $(\alpha,\beta)$ statically, we propose to adapt them dynamically during training so that the model approaches a desired false-negative to false-positive ratio. This leads to our proposed ATPO, which uses a feedback controller that dynamically adjusts the reward function during training. The overall training pipeline is illustrated in Figure~\ref{fig:method}. Training first performs SFT warmup, then the model is  optimized using ATPO, starting with the GRPO trainer with initialized Tversky coefficients. We instantiate ATPO with two variants of reward formulation. For brevity, we omit the GRPO group notation. The full algorithm is provided in Appendix~\ref{app:full_alg}.


\noindent\textbf{Category-Aware Adaptive Tversky Reward (C-ATR)} maintains separate controllers and adaptive Tversky coefficients $(\alpha_{t,c}, \beta_{t,c})$ for each category $c \in \mathcal{C}$.
At each training step $t$, GRPO rollouts produce predictions from which we compute batch-level counts of TP, FP, and FN:
\begin{equation}
TP_{t,c} = \sum_{i} \mathbf{1}[c \in P_i \cap G_i], \quad
FP_{t,c} = \sum_{i} \mathbf{1}[c \in P_i \setminus G_i], \quad
FN_{t,c} = \sum_{i} \mathbf{1}[c \in G_i \setminus P_i],
\end{equation}
where $P_i$ and $G_i$ denote the predicted and ground-truth label sets for sample $i$.
Following Step 3 in Figure~\ref{fig:method}, we maintain Exponential Moving Averages (EMA) of false positives and false negatives:
\begin{equation}
FP^{\text{ema}}_{t,c}
=
\lambda \, FP^{\text{ema}}_{t-1,c}
+
(1-\lambda)\,FP_{t,c}, \quad
FN^{\text{ema}}_{t,c}
=
\lambda \, FN^{\text{ema}}_{t-1,c}
+
(1-\lambda)\,FN_{t,c}
\end{equation}
where $\lambda \in [0,1)$ is a smoothing factor.
Using the EMA statistics, we update the estimate of the observed FN-to-FP ratio for category $c$ in Step 4: $r_{t,c} = 
(FN^{\text{ema}}_{t,c} + \varepsilon) / 
     (FP^{\text{ema}}_{t,c} + \varepsilon)$, 
where $\varepsilon>0$ ensures numerical stability. Then in Step 5, the ATPO controller for category $c$ measures the error by comparing the observed FN-to-FP ratio with user-given category-specific target ratio $r_c^*$ and updates a logit parameter $u_{t,c}$ that governs the relative weighting between false positives and false negatives:
\begin{equation}
e_{t,c}
=
\log r_{t,c}
-
\log r_c^*, \qquad u_{t+1,c}
=
u_{t,c}
+
\eta \, e_{t,c},
\label{eq:catr_controller}
\end{equation}
where $\eta>0$ is the step size.
We then compute the Tversky coefficients for the next training step that satisfy
\begin{equation}
\log \frac{\beta_{t+1,c}}{\alpha_{t+1,c}} = u_{t+1,c},
\qquad
\alpha_{t+1,c} + \beta_{t+1,c} = 2.
\end{equation}
The second constraint fixes the overall scale of the coefficients and recovers the Jaccard Index at the neutral operating point $\alpha_{t+1,c}=\beta_{t+1,c}=1$.
Solving the system yields
$
\alpha_{t+1,c}
=
2\bigl(1 - \sigma(u_{t+1,c})\bigr)$ and $\beta_{t+1,c}
=
2\sigma(u_{t+1,c})$, where $\sigma(\cdot)$ denotes the sigmoid function.
C-ATR uses micro-style aggregation across categories to calculate the reward:
\begin{equation}
R_t^{\text{C-ATR}}
=
\frac{\sum_{c} TP_{t,c}}
     {\sum_{c}
      \bigl(
        TP_{t,c}
        +
        \alpha_{t,c} FP_{t,c}
        +
        \beta_{t,c} FN_{t,c}
      \bigr)}
\label{eq:catr_reward}
\end{equation}
for training step $t$. By adapting $(\alpha_{t,c},\beta_{t,c})$ independently, C-ATR allows each category to converge toward its own target precision-recall operating point within a unified multi-label objective.

\noindent\textbf{Global Adaptive Tversky Reward (G-ATR)} is a special case of C-ATR that enforces a single global trade-off shared across all categories. \label{subsubsec:gatr}
Instead of maintaining per-category statistics, we first aggregate errors:
\begin{equation}
TP_t = \sum_{c} TP_{t,c},
\quad
FP_t = \sum_{c} FP_{t,c},
\quad
FN_t = \sum_{c} FN_{t,c}.
\end{equation}
We then compute global EMA statistics and the global FN-to-FP ratio and apply the same log-ratio control mechanism as in Equation~\ref{eq:catr_controller}, 
but with a single logit parameter $u_t$ and target ratio $r^*$.
The resulting adaptive coefficients $(\alpha_t,\beta_t)$ are shared across all categories, and the reward reduces to the standard global Tversky Index:
\begin{equation}
R_t^{\text{G-ATR}}
=
\frac{TP_t}
     {TP_t + \alpha_t FP_t + \beta_t FN_t}.
\end{equation}
Compared to C-ATR, G-ATR enforces a uniform precision-recall trade-off across all categories, whereas C-ATR enables category-specific operating points.

\section{Experiments}
\label{sec:exp}
In this section, we evaluate ATPO and the proposed Adaptive Tversky Reward (ATR) on two video safety benchmarks. We first compare our models against large general-purpose VLMs and GRPO with static reward designs. We then analyze ATR’s ability to control the precision-recall trade-off both globally and at the category level, and study the training dynamics of the adaptive controller. 

\subsection{Experiment Setup}
\label{subsec:exp_setup}



Models are evaluated on two public video safety benchmarks: SafeWatch-Bench~\cite{ICLR2025_beac6bfb} and XD-Violence~\cite{Wu2020not}. SafeWatch-Bench contains 199,604 training videos and 1,420 test videos across six unsafe categories. We use 11,036 full training videos under two minutes for SFT and GRPO, and evaluate on the 820-video SafeWatch-Bench-Real test set. For XD-Violence, we use 3,520 training videos under five minutes and evaluate on the full 800-video test set. Additional dataset statistics, including category co-occurrence, are provided in Appendix~\ref{appsub:data_stat}.

We adopt Qwen2.5-VL-7B-Instruct~\cite{bai2025qwen25vltechnicalreport} and Qwen3-VL-8B-Instruct~\cite{bai2025qwen3vltechnicalreport} as backbone models, both of which support direct video inputs via uniform frame sampling. For the two training stages, SFT is performed with LoRA~\cite{hu2022lora} using ground-truth safety labels as targets. GRPO training starts from the SFT one-epoch checkpoints and optimizes the reward formulations described in Section~\ref{subsec:grpo} and \ref{subsec:atr}, together with a small format reward. More implementation details are provided in Appendix~\ref{appsub:implement}.


Four standard multi-label evaluation metrics are reported~\cite{6471714,Tsoumakas2010,NIPS2015_85f007f8}: Jaccard Index, Micro-F1, Macro-F1, and Binary F1. Jaccard Index measures the intersection-over-union between predicted and ground-truth label sets, as defined in Section~\ref{subsec:grpo}. Micro-F1 computes F1 after aggregating true positives, false positives, and false negatives across categories, while Macro-F1 averages category-wise F1 scores. Binary F1 collapses all unsafe categories into a single positive class, following prior video safety detection works~\cite{huang2025vadr,liu2025guardreasonervl}.


\begin{table}[tb]
\small
  \caption{Main results on \textbf{SafeWatch-Bench-Real} and \textbf{XD-Violence}. We report multi-label metrics (Jaccard, Micro F1, Macro F1) and binary detection performance (Binary F1). Models are grouped into general-purpose VLMs, video reasoning VLMs, guardrail systems, and our ATPO-trained models with G-ATR and C-ATR. \textbf{Bold} and \underline{underlining} indicate the best and second-best results.  $^*$: further trained on the respective benchmark training sets. $^\dagger$: SafeWatch-8B uses full videos and clips from SafeWatch-Bench.}
  \label{tab:main}
  \centering
  \begin{tabular}{@{}llcccccccc@{}}
    \toprule
    \multirow{2}{*}{\#} &
    \multirow{2}{*}{Model} 
& \multicolumn{4}{c}{SafeWatch-Bench-Real} 
& \multicolumn{4}{c}{XD-Violence} \\

\cmidrule(lr){3-6} \cmidrule(lr){7-10}

& & Jaccard & Micro & Macro & Binary
& Jaccard & Micro & Macro & Binary \\
    \midrule
    \multicolumn{10}{c}{\textit{General-purpose VLMs}} \\
    1 & Qwen2.5VL-7B & 40.66 & 42.17 & 28.74 & 65.94 & 76.54 & 72.29 & 62.78 & 90.21 \\
    2 & Qwen2.5VL-32B & 41.66 & 43.60 & 32.11 & 64.63 & 77.69 & 73.28 & 62.71 & 91.85 \\
    3 & Qwen3VL-8B & 48.69 & 52.45 & 40.32 & 77.76 & 81.46 & 77.15 & 70.87 & 94.76 \\
    4 & Qwen3VL-235B-A22B & 55.85 & 62.10 & 56.83 & 80.29 & 82.53 & 78.03 & {\bf 72.29} & 95.36 \\
    5 & InternVL3.5-38B & 60.61 & 64.05 & 53.55 & 85.80 & 83.38 & 79.59 & 70.87 & 93.59 \\
    \multicolumn{10}{c}{\textit{Video reasoning VLMs}} \\
    6 & Video-R1 & 39.99 & 39.73 & 23.98 & 66.07 & 76.60 & 72.55 & 66.02 & 90.37 \\
    7 & VideoChat-R1 & 44.20 & 47.47 & 33.99 & 75.02 & 72.62 & 66.82 & 58.88 & 89.31 \\
    \multicolumn{10}{c}{\textit{Guardrails}} \\
    8 & GuardReasoner-VL & 19.90 & 1.17 & 0.98 & 1.22 & 35.34 & 3.09 & 2.78 & 4.50 \\
    9 & VAD-R1 & 47.78 & 50.24 & 43.90 & 80.95 & - & - & - & - \\
    10 & VAD-R1* & 46.36 & 51.44 & 48.93 & 81.13 & 74.12 & 68.60 & 68.81 & 89.69 \\
    11 & SafeWatch-8B$^\dagger$ & 70.69 & 73.15 & 71.37 & 89.04 & 79.84 & 77.19 & 69.65 & 86.80 \\
    \multicolumn{10}{c}{\textit{ATPO-trained (ours)}} \\
    12 & ATPO-G-7B & \underline{75.44} & \underline{77.25} & \underline{76.10} & 94.82 & {\bf 88.17} & {\bf 85.24} & \underline{71.95} & {\bf 96.69} \\
    13 & ATPO-C-7B & 75.15 & {\bf 77.42} & {\bf 76.61} & {\bf 95.25} & \underline{87.60} & \underline{84.51} & 71.24 & 95.59 \\
    14 & ATPO-G-8B & {\bf 75.50} & 77.08 & 75.25 & \underline{95.22} & 85.93 & 82.82 & 69.86 & 94.59 \\
    15 & ATPO-C-8B & 74.63 & 77.19 & 76.09 & 93.79 & 86.60 & 83.30 & 70.51 & \underline{95.75} \\

  \bottomrule
  \end{tabular}
\end{table}

\subsection{Multi-Label Video Safety Detection Performance}
\label{subsec:exp_main}
We evaluate several representative VLMs as baselines, including large general-purpose VLMs of different scales from the Qwen-VL and InternVL3.5 families. We also evaluate recent GRPO-trained video reasoning models, Video-R1~\cite{feng2025videor} and VideoChat-R1~\cite{yan2025videochatr}, which are built on Qwen2.5-VL-7B and trained for general video understanding tasks. Our guardrail baselines include GuardReasoner-VL~\cite{liu2025guardreasonervl}, VAD-R1~\cite{huang2025vadr}, and SafeWatch-8B~\cite{ICLR2025_beac6bfb}, a specialized multi-label video guardrail trained on SafeWatch-Bench.

Evaluation results for baseline models, prior work, and our models on SafeWatch-Bench-Real and XD-Violence are reported in Table~\ref{tab:main}. Our models are denoted as ATPO-G and ATPO-C, corresponding to ATPO trained with G-ATR and C-ATR, respectively, and are based on Qwen2.5-VL-7B and Qwen3-VL-8B backbones. The results reveal several key findings.

\noindent\textbf{Multi-VSD remains challenging.} Even very large VLMs (row~4 and row~5) still achieve only moderate Multi-VSD performance on SafeWatch-Bench-Real (Jaccard $<61$), despite relatively strong binary detection (Binary F1 $>85$). In contrast, most models obtain consistently higher scores on XD-Violence, suggesting XD-Violence is an easier benchmark under the evaluation metrics. Notably, models trained with GRPO for general video reasoning do not yield substantial gains. Video-R1 (row~6) and VideoChat-R1 (row~7) both build on the Qwen2.5-VL-7B backbone. Compared to the backbone's  results (row~1), Video-R1 shows no improvement (Jaccard $39.99$ \vs $40.66$), and VideoChat-R1 provides only a modest gain (Jaccard $44.20$ \vs $40.66$). This indicates that general video reasoning training alone is insufficient for Multi-VSD, and that dedicated in-domain training is likely necessary.

\noindent\textbf{ATPO improves Multi-VSD performance.} ATPO substantially improves Multi-VSD performance over the backbone models. For Qwen2.5-VL-7B, G-ATR and C-ATR (rows~12-13) improve SafeWatch-Bench-Real Jaccard from $40.66$ (row~1) to $75.44$/$75.15$ (a $\sim$+$35$ point gain). For Qwen3-VL-8B, G-ATR and C-ATR (rows~14-15) improve Jaccard from $48.69$ (row~3) to $75.50$/$74.63$. Overall, ATPO-trained models substantially outperform both larger general-purpose VLMs and video reasoning VLMs based on the same backbone model. 

\noindent\textbf{ATPO outperforms specialized guardrails.} SafeWatch-8B (row~11) provides a strong in-domain baseline, achieving the best non-ATPO results across all four metrics on SafeWatch-Bench-Real. Nevertheless, all four ATPO variants outperform SafeWatch-8B across all eight dataset-metric combinations. In particular, ATPO-G-7B achieves Jaccard scores of $75.44$ and $88.17$ on SafeWatch-Bench-Real and XD-Violence, compared with $70.69$ and $79.84$ for SafeWatch-8B, corresponding to gains of $4.75$ and $8.33$ points. In Appendix~\ref{app:multilabel} we also show that ATPO outperforms SafeWatch-8B and other training baselines on videos with multiple labels only.

Besides Multi-VSD, \textbf{ATPO-trained models also excel at binary classification.} All ATPO-trained models achieve Binary F1 $>93$ on both datasets (rows~12-15), indicating that optimizing for multi-label safety detection also leads to strong binary safety detection performance. Notably, the binary performance of these models is higher than guardrail models that are trained specifically for binary detection. Another finding from the results is that \textbf{G-ATR and C-ATR have comparable overall performance.} Across both backbones, G-ATR and C-ATR are very close (rows~12 \vs 13 and rows~14 \vs 15), with minor trade-offs across metrics. This suggests robustness to the choice of global \vs category-wise control in our ATPO training.

Overall, Table~\ref{tab:main} presents our best results using in-domain training with ATPO, showing substantial improvements over zero-shot general-purpose models. We further examine transfer across the SafeWatch-Bench and XD-Violence taxonomies, including adaptation with only $10\%$ of the target training data, in Appendix~\ref{app:taxonomy_transfer}. In the subsequent table, we show the relative contributions of our in-domain training techniques.

\subsection{Comparison of Static and Adaptive Reward Optimization}
\label{subsec:reward_comp}


Table~\ref{tab:reward} compares zero-shot evaluation, SFT, GRPO with static rewards (EM and STR), and ATPO with adaptive rewards (G-ATR and C-ATR), using Qwen2.5-VL-7B. We evaluate on both the real and AI-generated test subsets of SafeWatch-Bench~\cite{ICLR2025_beac6bfb}, referred to as Real and GenAI below.


\noindent\textbf{ATPO substantially improves over its initialization.} Starting from SFT-1ep, ATPO-G and ATPO-C improve Real Jaccard from $44.33$ to $75.15$, respectively, corresponding to gains of $31.11$ and $30.82$ points. GenAI Jaccard also increases from $66.34$ to $74.29$ and $73.35$, respectively. Both variants improve all four metrics on both subsets relative to their shared initialization.

\begin{table}[tb]
  \centering
  \small
  \setlength{\tabcolsep}{3pt}
  \caption{Training comparison on Qwen2.5-VL-7B. GRPO uses static rewards (Exact Match and STR), whereas ATPO-G and ATPO-C use adaptive rewards G-ATR and C-ATR, respectively. All RL methods start from the SFT-1ep checkpoint. Micro, Macro, and Binary denote F1 scores. \textbf{Bold} and \underline{underlining} indicate the best and second-best results per column.}
  \label{tab:reward}

  \begin{tabular}{@{}lcccccccc@{}}
    \toprule
    \multirow{2}{*}{Training}
      & \multicolumn{4}{c}{SafeWatch-Bench-Real}
      & \multicolumn{4}{c}{SafeWatch-Bench-GenAI} \\
    \cmidrule(lr){2-5} \cmidrule(lr){6-9}
      & Jaccard & Micro & Macro & Binary
      & Jaccard & Micro & Macro & Binary \\
    \midrule

    Zero-shot
      & 40.66 & 42.17 & 28.74 & 65.94
      & 47.18 & 43.30 & 27.18 & 75.54 \\

    SFT-1ep \textit{(RL initialization)}
      & 44.33 & 44.73 & 38.47 & 63.53
      & 66.34 & 71.40 & 67.90 & 90.20 \\

    SFT-3ep \textit{(extended SFT)}
      & 67.93 & 72.89 & 71.05 & 92.02
      & 62.32 & 70.49 & 68.11 & 87.86 \\

    \midrule
    \multicolumn{9}{l}{
      \textit{RL from the same SFT-1ep checkpoint}
    } \\
    \addlinespace[2pt]

    GRPO (EM)
      & 63.36 & 67.59 & 63.88 & 92.30
      & 62.23 & 68.84 & 64.79 & 89.48 \\

    GRPO (STR)
      & 70.09 & 72.27 & 63.88 & 93.97
      & 69.12 & 73.79 & 64.38 & 91.12 \\

    ATPO-G
      & \textbf{75.44} & \underline{77.25}
      & \underline{76.10} & \underline{94.82}
      & \textbf{74.29} & \textbf{76.24}
      & \textbf{74.46} & \underline{93.30} \\

    ATPO-C
      & \underline{75.15} & \textbf{77.42}
      & \textbf{76.61} & \textbf{95.25}
      & \underline{73.35} & \underline{74.30}
      & \underline{74.44} & \textbf{94.05} \\

    \bottomrule
  \end{tabular}
\end{table}


\noindent\textbf{Extended SFT trades off Real and GenAI performance.} AI-generated videos are less represented in the SafeWatch-Bench training set compared to real videos. Extending SFT from one to three epochs increases Real Jaccard from $44.33$ to $67.93$, but reduces GenAI Jaccard from $66.34$ to $62.32$. GenAI Micro and Binary F1 also decline. Thus, the gains from extended SFT do not hold across both video types. In contrast, both ATPO variants outperform SFT-3ep across all four metrics on both subsets. Cross-dataset evaluation on VHD11K-Video~\cite{yeh2024t2vs} shows a similar pattern for extended SFT and ATPO-C (Appendix~\ref{app:ood}).


\noindent\textbf{Adaptive rewards outperform static rewards.} Among the static rewards, STR achieves higher Jaccard than EM on both subsets, consistent with the benefit of graded feedback for partially correct predictions. Both ATPO variants further outperform EM and STR across all eight dataset-metric combinations. Relative to GRPO with STR, ATPO-G improves Jaccard by $5.35$ on Real ($70.09$ to $75.44$) and $5.17$ points on GenAI ($69.12$ to $74.29$). ATPO-C achieves corresponding gains of $5.06$ and $4.23$ points. These matched-RL comparisons support the contribution of adaptive reward design and suggest that adaptively rebalancing FN/FP during training provides a more informative learning signal than EM or STR (qualitative analysis in Appendix~\ref{app:qual}). Moreover, while SFT and static rewards impose a fixed operating point, ATR enables stable RL optimization with controllable error trade-offs, which we examine in the next section. Additional Qwen3-VL-8B results are provided in Appendix~\ref{app:reward_qwen3}.

\begin{table}[t]
\small
  \caption{Global precision-recall control on SafeWatch-Bench-Real. Comparison between Static Tversky Reward (STR) and Global Adaptive Tversky Reward (G-ATR) under varying control ratios. We report micro Precision (P), Recall (R), their ratio (P/R), and Error Type Ratio (ETR = FN/FP).}
  \label{tab:g_control}
  \centering
  \begin{tabular}{>{\columncolor{gray!15}}ccccc>{\columncolor{gray!15}}ccccc}
    \toprule
    
    \multicolumn{5}{c}{STR} & \multicolumn{5}{c}{G-ATR} \\
     \cmidrule(lr){1-5} \cmidrule(lr){6-10}
    $\alpha / \beta$ & Precision & Recall & P/R & ETR & $r^*$ & Precision & Recall & P/R & ETR \\
    \midrule
    1 / 5 & 81.24 & 73.21 & 1.11 & 1.58 & 0.2 & 66.33 & 86.55 & 0.77 & 0.31 \\
    1 / 2 & 79.11 & 69.88 & 1.13 & 1.63 & 0.5 & 76.18 & 80.71 & 0.94 & 0.76 \\
    1 / 1 & 71.93 & 72.62 & 0.99 & 0.97 & 1 & 75.74 & 78.81 & 0.96 & 0.84 \\
    2 / 1 & 75.03 & 71.55 & 1.05 & 1.20 & 2 & 75.72 & 69.05 & 1.10 & 1.40 \\
    5 / 1 & 73.48 & 77.50 & 0.95 & 0.80 & 5 & 78.56 & 67.62 & 1.16 & 1.75
 \\

  \bottomrule
  \end{tabular}
\end{table}

\subsection{Controlling Recall and Precision with ATR}
\label{subsec:gcontrol_exp}
Next, we investigate the question of whether ATR can be used to control the global precision-recall trade-off through the target FN/FP ratio $r^*$, and at the category level through per-category target ratio $r^*_c$.

\subsubsection{Global Control via G-ATR}

To quantify global control, we report micro Precision (P), micro Recall (R), and their ratio P/R in Table~\ref{tab:g_control}. We additionally report Error Type Ratio (ETR), defined as FN/FP, which directly measures the relative prevalence of type-II errors (false negatives) \vs type-I errors (false positives). For G-ATR, we vary the target ratio $r^*$: a smaller $r^*$ is expected to suppress FN and thus encourage a recall-dominant operating point, whereas a larger $r^*$ should suppress FP and shift the model toward a precision-dominant regime. For comparison, we evaluate the Static Tversky Reward (STR) defined in Equation~\ref{eq:tversky}, where the coefficients $\alpha$ and $\beta$ are fixed. Under STR, a smaller $\alpha / \beta$ places greater penalty on FN, while a larger $\alpha / \beta$ penalizes FP more. We make the following two observations from Table~\ref{tab:g_control}.

\textbf{G-ATR yields monotonic global control.} Across the full sweep of target ratios, G-ATR produces a clear and monotonic progression in the achieved precision-recall ratio. As $r^*$ increases, precision increases while recall decreases, and the achieved P/R rises steadily from $0.77 (r^*=0.2)$ to $1.16 (r^*=5)$. Notably, ETR follows the same intended direction: as $r^*$ grows, FN/FP increases monotonically from $0.31$ to $1.75$. Overall, varying $r^*$ smoothly shifts the model from a recall-favoring regime to a precision-favoring regime, suggesting that G-ATR functions as a stable feedback controller for global steering. The corresponding precision–recall curves are provided in Appendix~\ref{app:pr_curve}. In contrast, we find that \textbf{STR does not provide reliable control} under fixed FN and FP weightings. The achieved P/R under STR is non-monotonic and sometimes moves counter to the desired direction as $\alpha / \beta$ increases. The lack of consistency is further reflected by ETR, which does not exhibit a clean monotonic relationship with $\alpha / \beta$. This instability indicates that simply fixing the ratio $\frac{\alpha}{\beta}$ is insufficient for predictable global precision-recall steering during GRPO training. We additionally compare ATPO-G with post-hoc threshold tuning of SFT and GRPO (STR) in Appendix~\ref{app:threshold}.

\begin{table}[tb]
\small
  \caption{Category-level precision-recall control on SafeWatch-Bench-Real. Comparison between zero-shot large general-purpose VLMs (InternVL3.5-38B and Qwen3-VL-235B-A22B) and ATPO with C-ATR under symmetric and asymmetric target ratios for \textit{Misinformation} and \textit{Extreme}. We report category-wise Precision, Recall, their ratio P/R, and Error Type Ratio (ETR).}
  \label{tab:cat_control}
  \centering
  \begin{tabular}{lcccccccccc}
    \toprule
    \multirow{2}{*}{Model} & \multirow{2}{*}{$r^*_4$} & \multirow{2}{*}{$r^*_6$} & \multicolumn{4}{c}{C4 (\textit{Misinformation})} & \multicolumn{4}{c}{C6 (\textit{Extreme})} \\
    \cmidrule(lr){4-7} \cmidrule(lr){8-11}
     & & & Precision & Recall & P/R & ETR & Precision & Recall & P/R & ETR \\
    \midrule
    InternVL & - & - & 94.44 & 19.54 & 4.83 & 70.00 & 94.74 & 15.79 & 6.00 & 96.00 \\
    Qwen3-VL & - & - & 95.83 & 22.55 & 4.25 & 79.00 & 96.43 & 20.93 & 4.61 & 102.00 \\
    C-ATR & 1.0 & 1.0 & 55.28 & 66.67 & 0.83 & 0.62 & 80.31 & 79.07 & 1.02 & 1.08 \\
    C-ATR & 5.0 & 0.2 & 70.41 & 67.65 & 1.04 & 1.14 & 31.77 & 94.57 & 0.36 & 0.03 \\

  \bottomrule
  \end{tabular}
\end{table}

\subsubsection{Category-Level Control via C-ATR}

To evaluate category-level control, we select the two strongest general-purpose models from Table~\ref{tab:main}, InternVL3.5-38B and Qwen3-VL-235B-A22B, as reference models. We focus on C4 (\textit{Misinformation}) and C6 (\textit{Extreme}) because misinformation detection may prioritize higher precision to reduce over-censorship, whereas extreme content prioritizes higher recall to minimize harmful exposure. We report category-wise Precision, Recall, P/R, and ETR on SafeWatch-Bench-Real in Table~\ref{tab:cat_control}.

\noindent\textbf{Over-conservative behavior of large VLMs}. We find that large VLMs exhibit high-precision, low-recall "over-conservative" behavior. Both InternVL and Qwen3-VL show conservative behavior on both C4 and C6. They achieve very high precision ($>94$) but low recall ($<23$), resulting in disproportionately large P/R and very high ETR values. This indicates that, without task-specific alignment, even highly capable models tend not to flag these two evaluated categories, leading to many missed harmful videos. In comparison, \textbf{C-ATR enables controllable, category-level precision-recall trade-offs.} The controlled setting ($r^*_4=5.0, r^*_6=0.2$) demonstrates the desired asymmetric steering: C4 shifts toward higher precision (from $55.28$ to $70.41$), while C6 shifts toward high recall ($79.07$ to $94.57$). Overall, these results show that C-ATR can override the "overly conservative" zero-shot behavior and provide practical, category-aware operating points aligned with heterogeneous safety requirements. We report complete results for all six categories in
Appendix~\ref{app:category_full} and analyze controller
dynamics under different step sizes in
Appendix~\ref{app:hyp_analysis}. Practical guidance on
selecting error-ratio targets for deployment is provided
in Appendix~\ref{app:target_selection}.

\section{Conclusion}
In this paper, we introduced ATPO, an adaptive reward framework for multi-label harmful video detection that enables controllable precision–recall trade-offs during RL training. Experiments show that ATPO improves multi-label detection while maintaining strong binary safety performance. Furthermore, ATPO provides a flexible mechanism for aligning model behavior with different moderation policies by adjusting the desired precision–recall operating point.

\section{Acknowledgments}
Guangyu Yang is supported by Cambridge Commonwealth, European and International Trust for the undertaking of the PhD in Engineering at the University of Cambridge. This work was conducted during an internship at Xiaohongshu Inc.

Jingbiao Mei is supported by Cambridge Commonwealth, European and International Trust for the undertaking of the PhD in Engineering at the University of Cambridge.

Jinghong Chen is supported by the Warwick Postgraduate Studentship from Christ's College and the Huawei Hisilicon Studentship for the undertaking of the PhD in Engineering at the University of Cambridge.

Prof. Bill Byrne holds concurrent appointments as a Professor of Information Engineering at Cambridge University and as an Amazon Scholar. This publication describes work performed at Cambridge University and is not associated with Amazon.

We would also like to thank all the reviewers for their knowledgeable reviews.

\bibliographystyle{plainnat}
\bibliography{main}

@String(CVPR  = {IEEE Conf. Comput. Vis. Pattern Recog.})

@String(ICCV  = {Int. Conf. Comput. Vis.})

@String(ECCV  = {Eur. Conf. Comput. Vis.})

@String(NeurIPS = {Adv. Neural Inform. Process. Syst.})

@String(ICLR  = {Int. Conf. Learn. Represent.})

@String(AAAI  = {AAAI})

@String(CVPR  = {CVPR})

@String(ICCV  = {ICCV})

@String(ECCV  = {ECCV})

@String(NeurIPS = {NeurIPS})

@String(ICLR  = {ICLR})

@inproceedings{ICLR2025_beac6bfb,
 author = {Chen, Zhaorun and Pinto, Francesco and Pan, Minzhou and Li, Bo},
 booktitle = {International Conference on Learning Representations},
 editor = {Y. Yue and A. Garg and N. Peng and F. Sha and R. Yu},
 pages = {76566--76608},
 title = {SafeWatch: An Efficient Safety-Policy Following Video Guardrail Model with Transparent Explanations},
 url = {https://proceedings.iclr.cc/paper_files/paper/2025/file/beac6bfb7eac3d651307c16ac747df01-Paper-Conference.pdf},
 volume = {2025},
 year = {2025}
}

@inproceedings{Wu2020not,
title={Not only Look, but also Listen: Learning Multimodal Violence Detection under 
Weak Supervision},
author={Wu, Peng and Liu, jing and Shi, Yujia and Sun, Yujia and Shao, Fangtao 
and Wu, Zhaoyang and Yang, Zhiwei},
booktitle={European Conference on Computer Vision (ECCV)},
year={2020}
}

@misc{bai2025qwen25vltechnicalreport,
      title={Qwen2.5-VL Technical Report}, 
      author={Shuai Bai and Keqin Chen and Xuejing Liu and Jialin Wang and Wenbin Ge and Sibo Song and Kai Dang and Peng Wang and Shijie Wang and Jun Tang and Humen Zhong and Yuanzhi Zhu and Mingkun Yang and Zhaohai Li and Jianqiang Wan and Pengfei Wang and Wei Ding and Zheren Fu and Yiheng Xu and Jiabo Ye and Xi Zhang and Tianbao Xie and Zesen Cheng and Hang Zhang and Zhibo Yang and Haiyang Xu and Junyang Lin},
      year={2025},
      eprint={2502.13923},
      archivePrefix={arXiv},
      primaryClass={cs.CV},
      url={https://arxiv.org/abs/2502.13923}, 
}

@misc{bai2025qwen3vltechnicalreport,
      title={Qwen3-VL Technical Report}, 
      author={Shuai Bai and Yuxuan Cai and Ruizhe Chen and Keqin Chen and Xionghui Chen and Zesen Cheng and Lianghao Deng and Wei Ding and Chang Gao and Chunjiang Ge and Wenbin Ge and Zhifang Guo and Qidong Huang and Jie Huang and Fei Huang and Binyuan Hui and Shutong Jiang and Zhaohai Li and Mingsheng Li and Mei Li and Kaixin Li and Zicheng Lin and Junyang Lin and Xuejing Liu and Jiawei Liu and Chenglong Liu and Yang Liu and Dayiheng Liu and Shixuan Liu and Dunjie Lu and Ruilin Luo and Chenxu Lv and Rui Men and Lingchen Meng and Xuancheng Ren and Xingzhang Ren and Sibo Song and Yuchong Sun and Jun Tang and Jianhong Tu and Jianqiang Wan and Peng Wang and Pengfei Wang and Qiuyue Wang and Yuxuan Wang and Tianbao Xie and Yiheng Xu and Haiyang Xu and Jin Xu and Zhibo Yang and Mingkun Yang and Jianxin Yang and An Yang and Bowen Yu and Fei Zhang and Hang Zhang and Xi Zhang and Bo Zheng and Humen Zhong and Jingren Zhou and Fan Zhou and Jing Zhou and Yuanzhi Zhu and Ke Zhu},
      year={2025},
      eprint={2511.21631},
      archivePrefix={arXiv},
      primaryClass={cs.CV},
      url={https://arxiv.org/abs/2511.21631}, 
}

@inproceedings{
hu2022lora,
title={Lo{RA}: Low-Rank Adaptation of Large Language Models},
author={Edward J Hu and Yelong Shen and Phillip Wallis and Zeyuan Allen-Zhu and Yuanzhi Li and Shean Wang and Lu Wang and Weizhu Chen},
booktitle={International Conference on Learning Representations},
year={2022},
url={https://openreview.net/forum?id=nZeVKeeFYf9}
}

@inproceedings{10.5555/3600270.3602070,
author = {Wei, Jason and Wang, Xuezhi and Schuurmans, Dale and Bosma, Maarten and Ichter, Brian and Xia, Fei and Chi, Ed H. and Le, Quoc V. and Zhou, Denny},
title = {Chain-of-thought prompting elicits reasoning in large language models},
year = {2022},
isbn = {9781713871088},
publisher = {Curran Associates Inc.},
address = {Red Hook, NY, USA},
booktitle = {Proceedings of the 36th International Conference on Neural Information Processing Systems},
articleno = {1800},
numpages = {14},
location = {New Orleans, LA, USA},
series = {NIPS '22}
}

@inproceedings{zheng2024llamafactory,
  title={LlamaFactory: Unified Efficient Fine-Tuning of 100+ Language Models},
  author={Yaowei Zheng and Richong Zhang and Junhao Zhang and Yanhan Ye and Zheyan Luo and Zhangchi Feng and Yongqiang Ma},
  booktitle={Proceedings of the 62nd Annual Meeting of the Association for Computational Linguistics (Volume 3: System Demonstrations)},
  address={Bangkok, Thailand},
  publisher={Association for Computational Linguistics},
  year={2024},
  url={http://arxiv.org/abs/2403.13372}
}

@misc{zheng2025easyr1,
  title        = {EasyR1: An Efficient, Scalable, Multi-Modality RL Training Framework},
  author       = {Yaowei Zheng and Junting Lu and Shenzhi Wang and Zhangchi Feng and Dongdong Kuang and Yuwen Xiong},
  howpublished = {\url{https://github.com/hiyouga/EasyR1}},
  year         = {2025}
}

@inproceedings{
EXPOHM2026Mei,
title={Ex{PO}-{HM}: Learning to Explain-then-Detect for Hateful Meme Detection},
author={Jingbiao Mei and Mingsheng Sun and Jinghong Chen and Pengda Qin and Yuhong Li and Da Chen and Bill Byrne},
booktitle={The Fourteenth International Conference on Learning Representations},
year={2026},
url={https://openreview.net/forum?id=bEejbORUI5}
}

@inproceedings{Calabrese2024StrucuturedExplanationsHumanMod,
    title = "Explainability and Hate Speech: Structured Explanations Make Social Media Moderators Faster",
    author = {Calabrese, Agostina  and
      Neves, Leonardo  and
      Shah, Neil  and
      Bos, Maarten  and
      Ross, Bj{\"o}rn  and
      Lapata, Mirella  and
      Barbieri, Francesco},
    editor = "Ku, Lun-Wei  and
      Martins, Andre  and
      Srikumar, Vivek",
    booktitle = "Proceedings of the 62nd Annual Meeting of the Association for Computational Linguistics (Volume 2: Short Papers)",
    month = aug,
    year = "2024",
    address = "Bangkok, Thailand",
    publisher = "Association for Computational Linguistics",
    url = "https://aclanthology.org/2024.acl-short.38/",
    doi = "10.18653/v1/2024.acl-short.38",
    pages = "398--408"
}

@article{https://doi.org/10.1111/j.1469-8137.1912.tb05611.x,
author = {Jaccard, Paul},
title = {THE DISTRIBUTION OF THE FLORA IN THE ALPINE ZONE.},
journal = {New Phytologist},
volume = {11},
number = {2},
pages = {37-50},
doi = {https://doi.org/10.1111/j.1469-8137.1912.tb05611.x},
url = {https://nph.onlinelibrary.wiley.com/doi/abs/10.1111/j.1469-8137.1912.tb05611.x},
eprint = {https://nph.onlinelibrary.wiley.com/doi/pdf/10.1111/j.1469-8137.1912.tb05611.x},
year = {1912}
}

@article{https://doi.org/10.1037/0033-295X.84.4.327,
author = {Tversky, Amos},
title = {Features of similarity.},
journal = {Psychological Review},
volume = {84},
number = {4},
pages = {327-352},
doi = {https://doi.org/10.1037/0033-295X.84.4.327},
year = {1977}
}

@inproceedings{maaz-etal-2024-video,
    title = "Video-{C}hat{GPT}: Towards Detailed Video Understanding via Large Vision and Language Models",
    author = "Maaz, Muhammad  and
      Rasheed, Hanoona  and
      Khan, Salman  and
      Khan, Fahad",
    editor = "Ku, Lun-Wei  and
      Martins, Andre  and
      Srikumar, Vivek",
    booktitle = "Proceedings of the 62nd Annual Meeting of the Association for Computational Linguistics (Volume 1: Long Papers)",
    month = aug,
    year = "2024",
    address = "Bangkok, Thailand",
    publisher = "Association for Computational Linguistics",
    url = "https://aclanthology.org/2024.acl-long.679/",
    doi = "10.18653/v1/2024.acl-long.679",
    pages = "12585--12602"
}

@inproceedings{
feng2025videor,
title={Video-R1: Reinforcing Video Reasoning in {MLLM}s},
author={Kaituo Feng and Kaixiong Gong and Bohao Li and Zonghao Guo and Yibing Wang and Tianshuo Peng and Junfei Wu and Xiaoying Zhang and Benyou Wang and Xiangyu Yue},
booktitle={The Thirty-ninth Annual Conference on Neural Information Processing Systems},
year={2025},
url={https://openreview.net/forum?id=a2JTVVvcEl}
}

@inproceedings{
yan2025videochatr,
title={VideoChat-R1.5: Visual Test-Time Scaling to Reinforce Multimodal Reasoning by Iterative Perception},
author={Ziang Yan and Yinan He and Xinhao Li and Zhengrong Yue and Xiangyu Zeng and Yali Wang and Yu Qiao and Limin Wang and Yi Wang},
booktitle={The Thirty-ninth Annual Conference on Neural Information Processing Systems},
year={2025},
url={https://openreview.net/forum?id=oVDAfLuRie}
}

@misc{li2025videochatr1enhancingspatiotemporalperception,
      title={VideoChat-R1: Enhancing Spatio-Temporal Perception via Reinforcement Fine-Tuning}, 
      author={Xinhao Li and Ziang Yan and Desen Meng and Lu Dong and Xiangyu Zeng and Yinan He and Yali Wang and Yu Qiao and Yi Wang and Limin Wang},
      year={2025},
      eprint={2504.06958},
      archivePrefix={arXiv},
      primaryClass={cs.CV},
      url={https://arxiv.org/abs/2504.06958}, 
}

@inproceedings{lin-etal-2024-video,
    title = "Video-{LL}a{VA}: Learning United Visual Representation by Alignment Before Projection",
    author = "Lin, Bin  and
      Ye, Yang  and
      Zhu, Bin  and
      Cui, Jiaxi  and
      Ning, Munan  and
      Jin, Peng  and
      Yuan, Li",
    editor = "Al-Onaizan, Yaser  and
      Bansal, Mohit  and
      Chen, Yun-Nung",
    booktitle = "Proceedings of the 2024 Conference on Empirical Methods in Natural Language Processing",
    month = nov,
    year = "2024",
    address = "Miami, Florida, USA",
    publisher = "Association for Computational Linguistics",
    url = "https://aclanthology.org/2024.emnlp-main.342/",
    doi = "10.18653/v1/2024.emnlp-main.342",
    pages = "5971--5984"
}

@article{zhang2024holmes,
        title={Holmes-VAD: Towards Unbiased and Explainable Video Anomaly Detection via Multi-modal LLM},
        author={Zhang, Huaxin and Xu, Xiaohao and Wang, Xiang and Zuo, Jialong and Han, Chuchu and Huang, Xiaonan and Gao, Changxin and Wang, Yuehuan and Sang, Nong},
        journal={arXiv preprint arXiv:2406.12235},
        year={2024}
      }

@inproceedings{
huang2025vadr,
title={Vad-R1: Towards Video Anomaly Reasoning via Perception-to-Cognition Chain-of-Thought},
author={Chao Huang and Benfeng Wang and Wei Wang and Jie Wen and Chengliang Liu and Li Shen and Xiaochun Cao},
booktitle={The Thirty-ninth Annual Conference on Neural Information Processing Systems},
year={2025},
url={https://openreview.net/forum?id=6eLlczxbrp}
}

@inproceedings{atang2024hawk,
  title = {Hawk: Learning to Understand Open-World Video Anomalies},
  author = {Tang, Jiaqi and Lu, Hao and Wu, Ruizheng and Xu, Xiaogang and Ma, Ke and Fang, Cheng and Guo, Bin and Lu, Jiangbo and Chen, Qifeng and Chen, Ying-Cong},
  year = {2024},
  booktitle = {Neural Information Processing Systems (NeurIPS)}
}

@InProceedings{Ye_2025_CVPR,
    author    = {Ye, Muchao and Liu, Weiyang and He, Pan},
    title     = {VERA: Explainable Video Anomaly Detection via Verbalized Learning of Vision-Language Models},
    booktitle = {Proceedings of the Computer Vision and Pattern Recognition Conference (CVPR)},
    month     = {June},
    year      = {2025},
    pages     = {8679-8688}
}

@inbook{10.5555/3716662.3716795,
author = {Wei, Johnny Tian-Zheng and Zufall, Frederike and Jia, Robin},
title = {Operationalizing Content Moderation "Accuracy" in the Digital Services Act},
year = {2025},
publisher = {AAAI Press},
booktitle = {Proceedings of the 2024 AAAI/ACM Conference on AI, Ethics, and Society},
pages = {1527–1538},
numpages = {12}
}

@article{gongane2022detection,
  title   = {Detection and moderation of detrimental content on social media platforms: current status and future directions},
  author  = {Gongane, Vaishali U. and Munot, Mousami V. and Anuse, Alwin D.},
  journal = {Social Network Analysis and Mining},
  volume  = {12},
  number  = {1},
  pages   = {129},
  year    = {2022},
  publisher = {Springer},
  doi     = {10.1007/s13278-022-00951-3},
  issn    = {1869-5469}
}

@misc{deepseek-math,
  author = {Zhihong Shao and Peiyi Wang and Qihao Zhu and Runxin Xu and Junxiao Song and Mingchuan Zhang and Y.K. Li and Y. Wu and Daya Guo},
  title = {DeepSeekMath: Pushing the Limits of Mathematical Reasoning in Open Language Models},
  journal = {CoRR},
  volume = {abs/2402.03300},
  year = {2024},
  url = {https://arxiv.org/abs/2402.03300},
}

@InProceedings{Sultani_2018_CVPR,
author = {Sultani, Waqas and Chen, Chen and Shah, Mubarak},
title = {Real-World Anomaly Detection in Surveillance Videos},
booktitle = {The IEEE Conference on Computer Vision and Pattern Recognition (CVPR)},
month = {June},
year = {2018}
}

@INPROCEEDINGS{liu2018ano_pred, 
	author={W. Liu and W. Luo, D. Lian and S. Gao}, 
	booktitle={2018 IEEE Conference on Computer Vision and Pattern Recognition (CVPR)}, 
	title={Future Frame Prediction for Anomaly Detection -- A New Baseline}, 
	year={2018}
}

@article{Zanella2024HarnessingLL,
  title={Harnessing Large Language Models for Training-Free Video Anomaly Detection},
  author={Luca Zanella and Willi Menapace and Massimiliano Mancini and Yiming Wang and Elisa Ricci},
  journal={2024 IEEE/CVF Conference on Computer Vision and Pattern Recognition (CVPR)},
  year={2024},
  pages={18527-18536},
  url={https://api.semanticscholar.org/CorpusID:268819369}
}

@INPROCEEDINGS{11094660,
  author={Zhang, Huaxin and Xu, Xiaohao and Wang, Xiang and Zuo, Jialong and Huang, Xiaonan and Gao, Changxin and Zhang, Shanjun and Yu, Li and Sang, Nong},
  booktitle={2025 IEEE/CVF Conference on Computer Vision and Pattern Recognition (CVPR)}, 
  title={Holmes-VAU: Towards Long-term Video Anomaly Understanding at Any Granularity}, 
  year={2025},
  volume={},
  number={},
  pages={13843-13853},
  doi={10.1109/CVPR52734.2025.01292}}

@InProceedings{pmlr-v267-huang25ad,
  title = 	 {Ex-{VAD}: Explainable Fine-grained Video Anomaly Detection Based on Visual-Language Models},
  author =       {Huang, Chao and Shi, Yushu and Wen, Jie and Wang, Wei and Xu, Yong and Cao, Xiaochun},
  booktitle = 	 {Proceedings of the 42nd International Conference on Machine Learning},
  pages = 	 {25750--25761},
  year = 	 {2025},
  editor = 	 {Singh, Aarti and Fazel, Maryam and Hsu, Daniel and Lacoste-Julien, Simon and Berkenkamp, Felix and Maharaj, Tegan and Wagstaff, Kiri and Zhu, Jerry},
  volume = 	 {267},
  series = 	 {Proceedings of Machine Learning Research},
  month = 	 {13--19 Jul},
  publisher =    {PMLR},
  url = 	 {https://proceedings.mlr.press/v267/huang25ad.html}
}

@misc{wu2024openvocabularyvideoanomalydetection,
      title={Open-Vocabulary Video Anomaly Detection}, 
      author={Peng Wu and Xuerong Zhou and Guansong Pang and Yujia Sun and Jing Liu and Peng Wang and Yanning Zhang},
      year={2024},
      eprint={2311.07042},
      archivePrefix={arXiv},
      primaryClass={cs.CV},
      url={https://arxiv.org/abs/2311.07042}, 
}

@INPROCEEDINGS{11094336,
  author={Li, Fei and Liu, Wenxuan and Chen, Jingjing and Zhang, Ruixu and Wang, Yuran and Zhong, Xian and Wang, Zheng},
  booktitle={2025 IEEE/CVF Conference on Computer Vision and Pattern Recognition (CVPR)}, 
  title={Anomize: Better Open Vocabulary Video Anomaly Detection}, 
  year={2025},
  volume={},
  number={},
  pages={29203-29212},
  doi={10.1109/CVPR52734.2025.02719}}

@article{li2025reinforcement,
  title={Reinforcement Learning Tuning for VideoLLMs: Reward Design and Data Efficiency},
  author={Li, Hongyu and Han, Songhao and Liao, Yue and Luo, Junfeng and Gao, Jialin and Yan, Shuicheng and Liu, Si},
  journal={arXiv preprint arXiv:2506.01908},
  year={2025}
}

@inproceedings{zhang-etal-2025-improve,
    title = "Improve Vision Language Model Chain-of-thought Reasoning",
    author = "Zhang, Ruohong  and
      Zhang, Bowen  and
      Li, Yanghao  and
      Zhang, Haotian  and
      Sun, Zhiqing  and
      Gan, Zhe  and
      Yang, Yinfei  and
      Pang, Ruoming  and
      Yang, Yiming",
    editor = "Che, Wanxiang  and
      Nabende, Joyce  and
      Shutova, Ekaterina  and
      Pilehvar, Mohammad Taher",
    booktitle = "Proceedings of the 63rd Annual Meeting of the Association for Computational Linguistics (Volume 1: Long Papers)",
    month = jul,
    year = "2025",
    address = "Vienna, Austria",
    publisher = "Association for Computational Linguistics",
    url = "https://aclanthology.org/2025.acl-long.82/",
    doi = "10.18653/v1/2025.acl-long.82",
    pages = "1631--1662",
    ISBN = "979-8-89176-251-0"
}

@ARTICLE{6471714,
  author={Zhang, Min-Ling and Zhou, Zhi-Hua},
  journal={IEEE Transactions on Knowledge and Data Engineering}, 
  title={A Review on Multi-Label Learning Algorithms}, 
  year={2014},
  volume={26},
  number={8},
  pages={1819-1837},
  doi={10.1109/TKDE.2013.39}}

@Inbook{Tsoumakas2010,
author="Tsoumakas, Grigorios
and Katakis, Ioannis
and Vlahavas, Ioannis",
title="Mining Multi-label Data",
bookTitle="Data Mining and Knowledge Discovery Handbook",
year="2010",
publisher="Springer US",
address="Boston, MA",
pages="667--685",
isbn="978-0-387-09823-4",
doi="10.1007/978-0-387-09823-4_34",
url="https://doi.org/10.1007/978-0-387-09823-4_34"
}

@inproceedings{NIPS2015_85f007f8,
 author = {Koyejo, Oluwasanmi O and Natarajan, Nagarajan and Ravikumar, Pradeep K and Dhillon, Inderjit S},
 booktitle = {Advances in Neural Information Processing Systems},
 editor = {C. Cortes and N. Lawrence and D. Lee and M. Sugiyama and R. Garnett},
 pages = {},
 publisher = {Curran Associates, Inc.},
 title = {Consistent Multilabel Classification},
 url = {https://proceedings.neurips.cc/paper_files/paper/2015/file/85f007f8c50dd25f5a45fca73cad64bd-Paper.pdf},
 volume = {28},
 year = {2015}
}

@inproceedings{DBLP:conf/iclr/LoshchilovH19,
  author       = {Ilya Loshchilov and
                  Frank Hutter},
  title        = {Decoupled Weight Decay Regularization},
  booktitle    = {7th International Conference on Learning Representations, {ICLR} 2019,
                  New Orleans, LA, USA, May 6-9, 2019},
  publisher    = {OpenReview.net},
  year         = {2019},
  url          = {https://openreview.net/forum?id=Bkg6RiCqY7},
  bibsource    = {dblp computer science bibliography, https://dblp.org}
}

@inproceedings{
liu2025guardreasonervl,
title={GuardReasoner-{VL}: Safeguarding {VLM}s via Reinforced Reasoning},
author={Yue Liu and Shengfang Zhai and Mingzhe Du and Yulin Chen and Tri Cao and Hongcheng Gao and Cheng Wang and Xinfeng Li and Kun Wang and Junfeng Fang and Jiaheng Zhang and Bryan Hooi},
booktitle={The Thirty-ninth Annual Conference on Neural Information Processing Systems},
year={2025},
url={https://openreview.net/forum?id=Ku3XdvO88g}
}

@inproceedings{yeh2024t2vs,
 author={Chen Yeh and You-Ming Chang and Wei-Chen Chiu and Ning Yu},
 booktitle = {Advances in Neural Information Processing Systems},
 title={T2Vs Meet VLMs: A Scalable Multimodal Dataset for Visual Harmfulness Recognition},
 year = {2024}
}

@article{luo2017revisit,
  title={A revisit of sparse coding based anomaly detection in stacked rnn framework},
  author={Luo, Weixin and Liu, Wen and Gao, Shenghua},
  journal={ICCV, Oct},
  volume={1},
  number={2},
  pages={3},
  year={2017}
}

@article{luo2019video,
  title={Video Anomaly Detection With Sparse Coding Inspired Deep Neural Networks},
  author={Luo, Weixin and Liu, Wen and Lian, Dongze and Tang, Jinhui and Duan, Lixin and Peng, Xi and Gao, Shenghua},
  journal={IEEE Transactions on Pattern Analysis and Machine Intelligence},
  year={2019},
  publisher={IEEE}
}

@inproceedings{park2020learning,
  title={Learning Memory-guided Normality for Anomaly Detection},
  author={Park, Hyunjong and Noh, Jongyoun and Ham, Bumsub},
  booktitle={Proceedings of the IEEE/CVF Conference on Computer Vision and Pattern Recognition},
  pages={14372--14381},
  year={2020}
}

@misc{li2025tempsampr1effectivetemporalsampling,
      title={TempSamp-R1: Effective Temporal Sampling with Reinforcement Fine-Tuning for Video LLMs}, 
      author={Yunheng Li and Jing Cheng and Shaoyong Jia and Hangyi Kuang and Shaohui Jiao and Qibin Hou and Ming-Ming Cheng},
      year={2025},
      eprint={2509.18056},
      archivePrefix={arXiv},
      primaryClass={cs.CV},
      url={https://arxiv.org/abs/2509.18056}, 
}

@inproceedings{3737916.3739955,
author = {Miao, Yibo and Zhu, Yifan and Yu, Lijia and Zhu, Jun and Gao, Xiao-Shan and Dong, Yinpeng},
title = {T2VSafetyBench: evaluating the safety of text-to-video generative models},
year = {2024},
isbn = {9798331314385},
publisher = {Curran Associates Inc.},
address = {Red Hook, NY, USA},
booktitle = {Proceedings of the 38th International Conference on Neural Information Processing Systems},
articleno = {2039},
numpages = {15},
location = {Vancouver, BC, Canada},
series = {NIPS '24}
}

@inproceedings{yoon2025safree,
        title={SAFREE: Training-Free and Adaptive Guard for Safe Text-to-Image And Video Generation}, 
        author={Jaehong Yoon and Shoubin Yu and Vaidehi Patil and Huaxiu Yao and Mohit Bansal},
        booktitle={International Conference on Learning Representations},
        year={2025},
}

@misc{sun2026evaluationdefenseadvancingsafety,
  title={From Evaluation to Defense: Advancing Safety in Video Large Language Models},
  author={Yiwei Sun and Peiqi Jiang and Chuanbin Liu and Luohao Lin and Zhiying Lu and Hongtao Xie},
  year={2026},
  eprint={2505.16643},
  archivePrefix={arXiv},
  primaryClass={cs.CV},
  url={https://arxiv.org/abs/2505.16643},
}

@inproceedings{
wang2026safevid,
title={SafeVid: Toward Safety Aligned Video Large Multimodal Models},
author={Yixu Wang and Jiaxin Song and Yifeng Gao and Xin Wang and Yang Yao and Yan Teng and Xingjun Ma and Yingchun Wang and Yu-Gang Jiang},
booktitle={The Thirty-ninth Annual Conference on Neural Information Processing Systems Datasets and Benchmarks Track},
year={2026},
url={https://openreview.net/forum?id=SeNFo7JGly}
}

@inproceedings{lu-etal-2025-sea,
    title = "{SEA}: Low-Resource Safety Alignment for Multimodal Large Language Models via Synthetic Embeddings",
    author = "Lu, Weikai  and
      Peng, Hao  and
      Zhuang, Huiping  and
      Chen, Cen  and
      Zeng, Ziqian",
    editor = "Che, Wanxiang  and
      Nabende, Joyce  and
      Shutova, Ekaterina  and
      Pilehvar, Mohammad Taher",
    booktitle = "Proceedings of the 63rd Annual Meeting of the Association for Computational Linguistics (Volume 1: Long Papers)",
    month = jul,
    year = "2025",
    address = "Vienna, Austria",
    publisher = "Association for Computational Linguistics",
    url = "https://aclanthology.org/2025.acl-long.1212/",
    doi = "10.18653/v1/2025.acl-long.1212",
    pages = "24894--24913",
    ISBN = "979-8-89176-251-0"
}

@inproceedings{
hu2025videojail,
title={VideoJail: Exploiting Video-Modality Vulnerabilities for Jailbreak Attacks on Multimodal Large Language Models},
author={Wenbo Hu and Shishen Gu and Youze Wang and Richang Hong},
booktitle={ICLR 2025 Workshop on Building Trust in Language Models and Applications},
year={2025},
url={https://openreview.net/forum?id=fSAIDcPduZ}
}

@INPROCEEDINGS {11092973,
author = { Liu, Haitong and Gao, Kuofeng and Bai, Yang and Li, Jinmin and Shan, Jinxiao and Dai, Tao and Xia, Shu-Tao },
booktitle = { 2025 IEEE/CVF Conference on Computer Vision and Pattern Recognition (CVPR) },
title = {{ Protecting Your Video Content: Disrupting Automated Video-based LLM Annotations }},
year = {2025},
volume = {},
ISSN = {},
pages = {24056-24065},
doi = {10.1109/CVPR52734.2025.02240},
url = {https://doi.ieeecomputersociety.org/10.1109/CVPR52734.2025.02240},
publisher = {IEEE Computer Society},
address = {Los Alamitos, CA, USA},
month =Jun}

@inproceedings{3666122.3668460,
author = {Rafailov, Rafael and Sharma, Archit and Mitchell, Eric and Ermon, Stefano and Manning, Christopher D. and Finn, Chelsea},
title = {Direct preference optimization: your language model is secretly a reward model},
year = {2023},
publisher = {Curran Associates Inc.},
address = {Red Hook, NY, USA},
booktitle = {Proceedings of the 37th International Conference on Neural Information Processing Systems},
articleno = {2338},
numpages = {14},
location = {New Orleans, LA, USA},
series = {NIPS '23}
}


\appendix

\section{Limitations}
\label{app:limit}
Our evaluation follows the category definitions and annotations provided by existing benchmarks (e.g., SafeWatch-Bench and XD-Violence), which reflect commonly adopted operational definitions of harmful content. However, the boundaries of categories such as misinformation, violence, or extremism may vary across cultural, legal, and platform-specific contexts. As a result, model behavior may differ under alternative definitions or policies. While ATPO provides a mechanism to adjust error trade-offs, adapting to different definitions may require dataset-specific annotation and validation.

ATPO enables controllable precision-recall trade-offs, but selecting appropriate target ratios requires application-specific tuning and validation. In particular, different operating points may lead to over-moderation (false positives) or missed harmful content (false negatives). Therefore, ATPO is best used as an assistive component in moderation pipelines to support human decision-making rather than serving as a fully automated standalone system.

\begin{algorithm}[th!]
\caption{ATPO with Category-Aware Adaptive Tversky Reward (C-ATR)}
\label{alg:catr}
\small
\begin{algorithmic}[1]
\REQUIRE Training set $\mathcal{D}=\{(x, v_i, G_i)\}_{i=1}^{|\mathcal{D}|}$, where $x$ is the detection instruction, $v_i$ is the input video and $G_i$ is the ground-truth label set, initial policy $\pi_\theta$, smoothing factor $\lambda$, step size $\eta$, stability constant $\varepsilon$, target ratios $\{r_c^*\}_{c\in\mathcal{C}}$, group size $N$
\ENSURE Trained policy $\pi_\theta$

\STATE Perform supervised fine-tuning (SFT) on $\mathcal{D}$ to initialize $\pi_{\theta_0}$

\FOR{each category $c\in\mathcal{C}$}
    \STATE Initialize controller state $u_{0,c}\gets 0$, EMA statistics $FP^{\mathrm{ema}}_{0,c}\gets 0$, $FN^{\mathrm{ema}}_{0,c}\gets 0$, and coefficients $\alpha_{0,c}\gets 1$, $\beta_{0,c}\gets 1$
\ENDFOR

\FOR{training step $t=0$ to $T-1$}
    \STATE Sample minibatch $\mathcal{B}=\{(x,v_i,G_i)\}_{i=1}^{M}$ from $\mathcal{D}$

    \FOR{each sample $(x,v_i,G_i)\in\mathcal{B}$}
        \STATE Sample a group of $N$ outputs $\{y_{i,j}\}_{j=1}^{N}\sim \pi_{\theta_t}(\cdot \mid x,v_i)$
        \STATE Extract predicted label sets $\{P_{i,j}\}_{j=1}^{N}$ from $\{y_{i,j}\}_{j=1}^{N}$

        \FOR{each output $j=1$ to $N$}
            \FOR{each category $c\in\mathcal{C}$}
                \STATE Compute $TP_{t, c, i, j} \gets \mathbf{1}[c\in P_{i,j}\cap G_i]$, $FP_{t, c, i, j} \gets \mathbf{1}[c\in P_{i,j}\setminus G_i]$, $FN_{t, c, i, j} \gets \mathbf{1}[c\in G_i\setminus P_{i,j}]$
            \ENDFOR
            \STATE Compute reward using current coefficients $\{\alpha_{t,c},\beta_{t,c}\}_{c\in\mathcal{C}}$\\
            $
            R_{t,i,j}\gets \frac{\sum_c TP_{t,c,i,j}}{\sum_c (TP_{t,c,i,j} + \alpha_{t,c}FP_{t,c,i,j} + \beta_{t,c}FN_{t,c,i,j})}
            $
        \ENDFOR

        \STATE Compute group-normalized advantages
    \ENDFOR

    \STATE Update policy from $\pi_{\theta_t}$ to $\pi_{\theta_{t+1}}$ with GRPO
    \FOR{each category $c\in\mathcal{C}$}
        \STATE Update EMA statistics\\
        $
        FP_{t,c} \gets \sum_{i=1}^{M}\sum_{j=1}^{N}FP_{t,c,i,j},\quad  FP^{\mathrm{ema}}_{t+1,c} \gets
        \lambda FP^{\mathrm{ema}}_{t,c} + (1-\lambda)FP_{t,c}
        $\\
        $
        FN_{t,c} \gets \sum_{i=1}^{M}\sum_{j=1}^{N}FN_{t,c,i,j},\quad
        FN^{\mathrm{ema}}_{t+1,c} \gets
        \lambda FN^{\mathrm{ema}}_{t,c} + (1-\lambda)FN_{t,c}
        $

        \STATE Compute observed ratio
        $r_{t,c} \gets \frac{FN^{\mathrm{ema}}_{t+1,c}+\varepsilon}{FP^{\mathrm{ema}}_{t+1,c}+\varepsilon}$

        \STATE Compute control error
        $
        e_{t,c} \gets \log r_{t,c}-\log r_c^*
        $

        \STATE Update controller
        $
        u_{t+1,c} \gets u_{t,c}+\eta e_{t,c}
        $
        \STATE Update next-step coefficients
        $
        \alpha_{t+1,c}\gets2\bigl(1-\sigma(u_{t+1,c})\bigr), \quad
        \beta_{t+1,c}\gets2\sigma(u_{t+1,c})
        $
    \ENDFOR
\ENDFOR

\RETURN $\pi_{\theta_T}$
\end{algorithmic}
\end{algorithm}

\section{Full Algorithm}
\label{app:full_alg}
In Section~\ref{subsec:atr}, we present the training pipeline of Adaptive Tversky Policy Optimization (ATPO). Here we provide the full training procedure of ATPO instantiated with the Category-Aware Adaptive Tversky Reward (C-ATR) in Algorithm~\ref{alg:catr}. The training begins with Supervised Fine-Tuning (SFT) to initialize the Vision-Language Model (VLM) with harmful category definitions. ATPO then iteratively performs GRPO rollouts and policy updates~\cite{deepseek-math}, followed by EMA updates, ratio estimation, and controller updates that adapt the Tversky coefficients~\cite{https://doi.org/10.1037/0033-295X.84.4.327} toward the target ratios.

\section{Additional Experiment Setup Details}
\label{app:exp_setup}
In this section, we provide additional details on the experimental setup for Section~\ref{subsec:exp_setup}.
\subsection{Statistics of Datasets}
\label{appsub:data_stat}
In Section~\ref{subsec:exp_setup}, we introduce the two test sets used for evaluation: SafeWatch-Bench-Real~\cite{ICLR2025_beac6bfb} and XD-Violence~\cite{Wu2020not}. Here we present additional dataset statistics to illustrate the characteristics of these datasets and the challenges they pose for Multi-label Video Safety Detection (Multi-VSD). Table~\ref{tab:label_dist} shows the distribution of samples with each harmful category and multi-label samples among all unsafe samples. We can see that the proportion of multi-label videos is substantially higher in SafeWatch-Bench-Real (25.52\%) than in XD-Violence (9.40\%). This indicates that SafeWatch more frequently contains videos exhibiting multiple harmful categories simultaneously, which increases prediction ambiguity and helps explain the greater difficulty observed for SafeWatch in Section~\ref{subsec:exp_main}. This highlights the importance of methods that can reliably handle multi-label predictions. In addition, XD-Violence  exhibits notable category imbalance, with the Abuse category accounting for only 2.2\% of unsafe samples. Such imbalance could  make it difficult for models to maintain consistent precision–recall trade-offs across categories.

\begin{figure}[tb]                                                              
      \centering                                                               
      \begin{subfigure}{0.48\textwidth}                                        
          \centering
          \includegraphics[width=\textwidth]{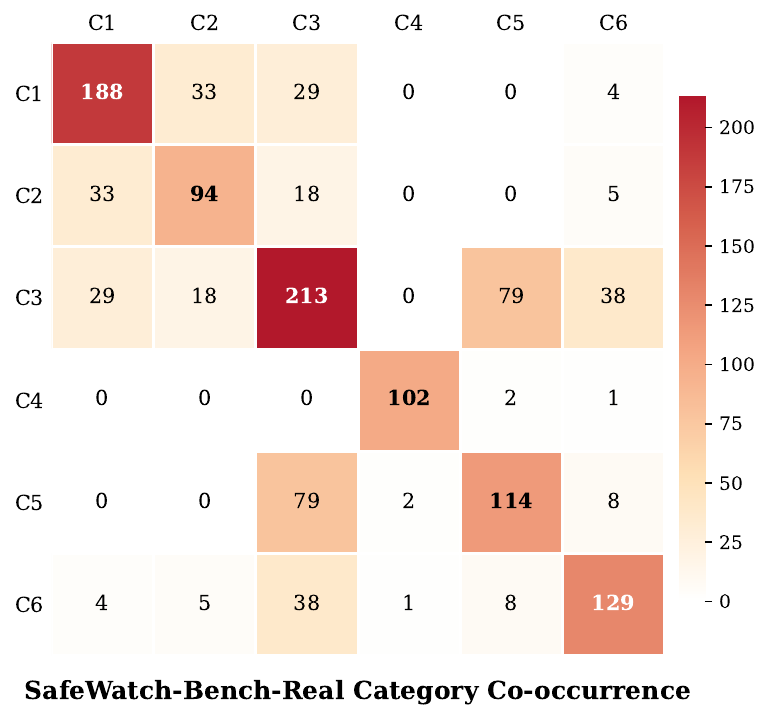}
      \end{subfigure}
      \hfill
      \begin{subfigure}{0.48\textwidth}
          \centering
          \includegraphics[width=\textwidth]{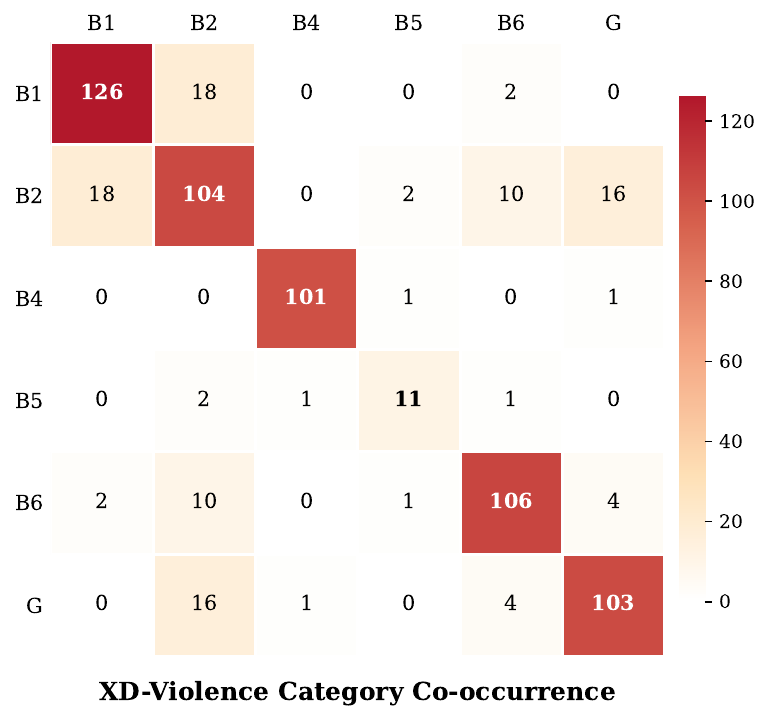}
      \end{subfigure}
      \caption{Category co-occurrence matrices for unsafe videos in the test sets of SafeWatch-Bench-Real (left) and XD-Violence (right), showing the number of videos where each pair of categories appears together.}
      \label{fig:cooccurrence}
  \end{figure}

Figure~\ref{fig:cooccurrence} further illustrates the relationships between categories by showing the category co-occurrence matrices for unsafe samples in the two test sets. We can see that several categories frequently co-occur in SafeWatch-Bench-Real, such as C3 (\textit{Violence}) with C5 (\textit{Illegal}) and C6 (\textit{Extreme}), whereas category overlap in XD-Violence is generally more limited. These patterns highlight the inherently multi-label nature of harmful videos, which motivates the development of ATPO for Multi-VSD.

  \begin{table}[tb]                                                                                                                                                                                       
  \centering                                                                                                                                                                                             
  \caption{Label distribution of unsafe categories in the SafeWatch-Bench-Real test set (650 harmful videos) and XD-Violence test set (500 violent videos). Category identifiers follow the original dataset annotations. Since the task is multi-label, videos may belong to multiple categories, so counts do not sum to the total number of videos.}
  \label{tab:label_dist}                                                                                                                                                                       
  \begin{tabular}{lcc}                                                                                                                                                                                   
  \toprule
  Category & Count & Percentage (\%) \\
  \midrule
  \multicolumn{3}{c}{SafeWatch-Bench-Real} \\
  \midrule
  C1 (Sexual)          & 188 & 28.92 \\
  C2 (Abuse)  &  94 & 14.46 \\
  C3 (Violence) & 213 & 32.77 \\
  C4 (Misinformation) & 102 & 15.69 \\
  C5 (Illegal)      & 114 & 17.54 \\
  C6 (Extreme)         & 129 & 19.85 \\
  Multi-label & 166 & 25.52 \\
  \midrule
  \multicolumn{3}{c}{XD-Violence} \\
  \midrule
  B1 (Fighting)      & 126 & 25.20 \\
  B2 (Shooting)      & 104 & 20.80 \\
  B4 (Riot)          & 101 & 20.20 \\
  B5 (Abuse)         &  11 &  2.20 \\
  B6 (Car Accident)  & 106 & 21.20 \\
  G (Explosion)      & 103 & 20.60 \\
  Multi-label & 47 & 9.40 \\
  \bottomrule
  \end{tabular}
  \end{table}

\subsection{Implementation Details}
\label{appsub:implement}
\begin{figure}[tb]                                                          
      \centering
          \centering
          \includegraphics[width=0.99\textwidth]{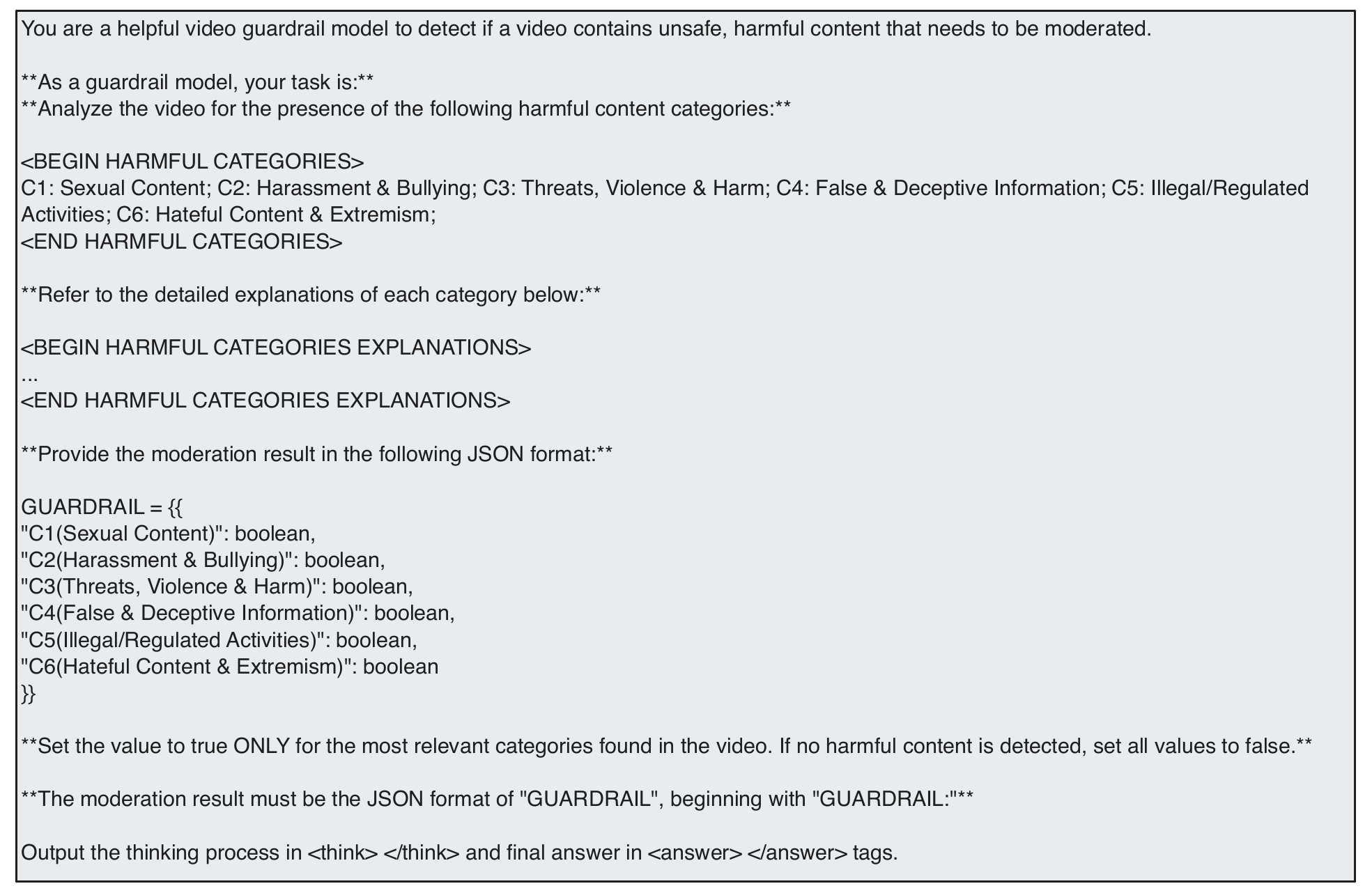}
      \caption{Prompt template used during ATPO training and inference on SafeWatch-Bench. Detailed descriptions of harmful categories are omitted for brevity.}
      \label{fig:prompt}
  \end{figure}

\noindent\textbf{Video Processing.} Videos are processed using the official preprocessing pipeline \texttt{qwen-vl-utils}~\cite{bai2025qwen25vltechnicalreport,bai2025qwen3vltechnicalreport}. Frames are sampled at 1 frame per second (FPS) without enforcing a fixed number of frames per video, so the sequence length varies with video duration. Frames are sampled uniformly across the video. Each frame is resized using dynamic resolution while preserving the original aspect ratio. 

\noindent\textbf{Prompt Format.}
The model receives a task instruction describing the video safety detection objective together with the category definitions. The prompt instructs the model to provide reasoning and output the predicted safety categories in a structured format following the original prompt used in SafeWatch-Bench~\cite{ICLR2025_beac6bfb}. The exact prompt template used during ATPO training and inference on SafeWatch-Bench is provided in Figure~\ref{fig:prompt}.

\noindent\textbf{Supervised Fine-Tuning (SFT).}
Training begins with an SFT stage to align the base VLM with the Multi-label Video Safety Detection (Multi-VSD) task. The SFT stage is trained for 1 epoch on the training split of each dataset using the standard language modeling objective. Additional hyperparameters follow common VLM fine-tuning practices, including the optimizer AdamW~\cite{DBLP:conf/iclr/LoshchilovH19}, learning rate $=10^{-5}$, effective batch size $=8$, and maximum sequence length $=24576$. The LoRA rank is set to 64 and the scaling factor is set to 128.

\noindent\textbf{Reinforcement Learning Training.}
After SFT, we further optimize the model using the proposed ATPO framework. The RL stage runs for 3 epochs. For each input sample, the model generates a group of responses to estimate policy advantages. The key RL hyperparameters are: group size $=8$, sampling temperature $=1.0$, top-$p=1.0$, and KL penalty coefficient $\beta_{\mathrm{KL}} = 10^{-2}$. The reference policy is the SFT-initialized model. For the ATPO controller, we use controller step size $\eta = 0.01$, EMA smoothing factor $\lambda = 0.95$, and stability constant $\varepsilon = 10^{-8}$. These parameters are shared across both the global and category-level variants of ATR.

\noindent\textbf{Training Infrastructure.}
All experiments are conducted on four NVIDIA H800 80GB GPUs. Training is performed using BF16 mixed-precision. The following libraries/packages were used for implementing the experiments: \texttt{LLaMA-Factory}~\cite{zheng2024llamafactory}, a modified version of \texttt{EasyR1}~\cite{zheng2025easyr1}, \texttt{PyTorch 2.8.0}, \texttt{vLLM 0.11.0}, \texttt{Transformers 4.57.3}, and \texttt{CUDA 12.8}. On average, RL training for three epochs takes approximately 20 hours on our GPU cluster.


\subsection{Prior Work}

We compare ATPO with recent approaches exploring Reinforcement Learning (RL) and reasoning capabilities in VLMs for video understanding and safety detection. Here we provide a more detailed description of those approaches that we evaluate in Section~\ref{subsec:exp_main}.

\noindent\textbf{GuardReasoner-VL.}
GuardReasoner-VL~\cite{liu2025guardreasonervl} proposes a reasoning-based VLM guardrail model for multi-modal safety moderation. The model is trained to generate an explicit reasoning process before producing moderation decisions. The model is first initialized with reasoning SFT and then further optimized with online GRPO using a length-aware safety reward. Although the original GuardReasoner-VL is trained and evaluated on text, image, and text-image data rather than video, its backbone model, Qwen2.5-VL-7B-Instruct, supports direct video inputs. Therefore, we evaluate GuardReasoner-VL on video inputs in our experiments for comparison with other models.

\noindent\textbf{VAD-R1.}
VAD-R1~\cite{huang2025vadr} proposes a reinforcement-learning-based framework for Video Anomaly Reasoning (VAR) that encourages VLMs to perform structured reasoning when analyzing anomalous events in videos. The model is trained using a two-stage pipeline: after initial supervised alignment, the model is optimized using Anomaly Verification Augmented GRPO (AVA-GRPO). AVA-GRPO extends GRPO with an anomaly verification reward that re-evaluates predictions by performing reasoning on temporally trimmed versions of the video. In our experiments, we further train VAD-R1 on the training set of SafeWatch-Bench using AVA-GRPO and report the evaluation results in Section~\ref{subsec:exp_main}.

\noindent\textbf{Video-R1 and VideoChat-R1.}
Video-R1~\cite{feng2025videor} extends the R1-style rule-based RL paradigm to video understanding by introducing Temporal Group Relative Policy Optimization (T-GRPO), a variant of GRPO designed to encourage general temporal reasoning over video frames. T-GRPO contrasts model performance on temporally ordered versus randomly shuffled frame sequences. 
VideoChat-R1~\cite{li2025videochatr1enhancingspatiotemporalperception} investigates reinforcement fine-tuning of video VLMs using GRPO with task-specific spatio-temporal rewards. The training incorporates reward functions tailored to several video perception tasks, including temporal grounding and video question answering, enabling the model to improve spatio-temporal perception while maintaining general dialogue capabilities.

\noindent\textbf{SafeWatch-8B.}
SafeWatch-8B~\cite{ICLR2025_beac6bfb} is a video guardrail model built on InternVL2-8B that produces multi-label safety decisions and content-specific explanation conditioned on safety policies. It introduces Parallel Equivalent Policy Encoding (PEPE) to encode policies independently and Policy-Aware Adaptive Pruning (PAP) to retain policy-relevant visual tokens. Its training pipeline comprises multi-task supervised fine-tuning, guardrail-specific fine-tuning with pruning on SafeWatch-Bench, and Direct Preference Optimization (DPO)~\cite{3666122.3668460} to improve explanations and reduce false positives. We include SafeWatch-8B as a specialized multi-label video guardrail baseline in Table~\ref{tab:main}.

All models are evaluated at 1 fps and the same inference prompt to ensure fair comparison.

\begin{figure}[t]                                                          
      \centering                                                            
          \centering
          \includegraphics[width=0.6\textwidth]{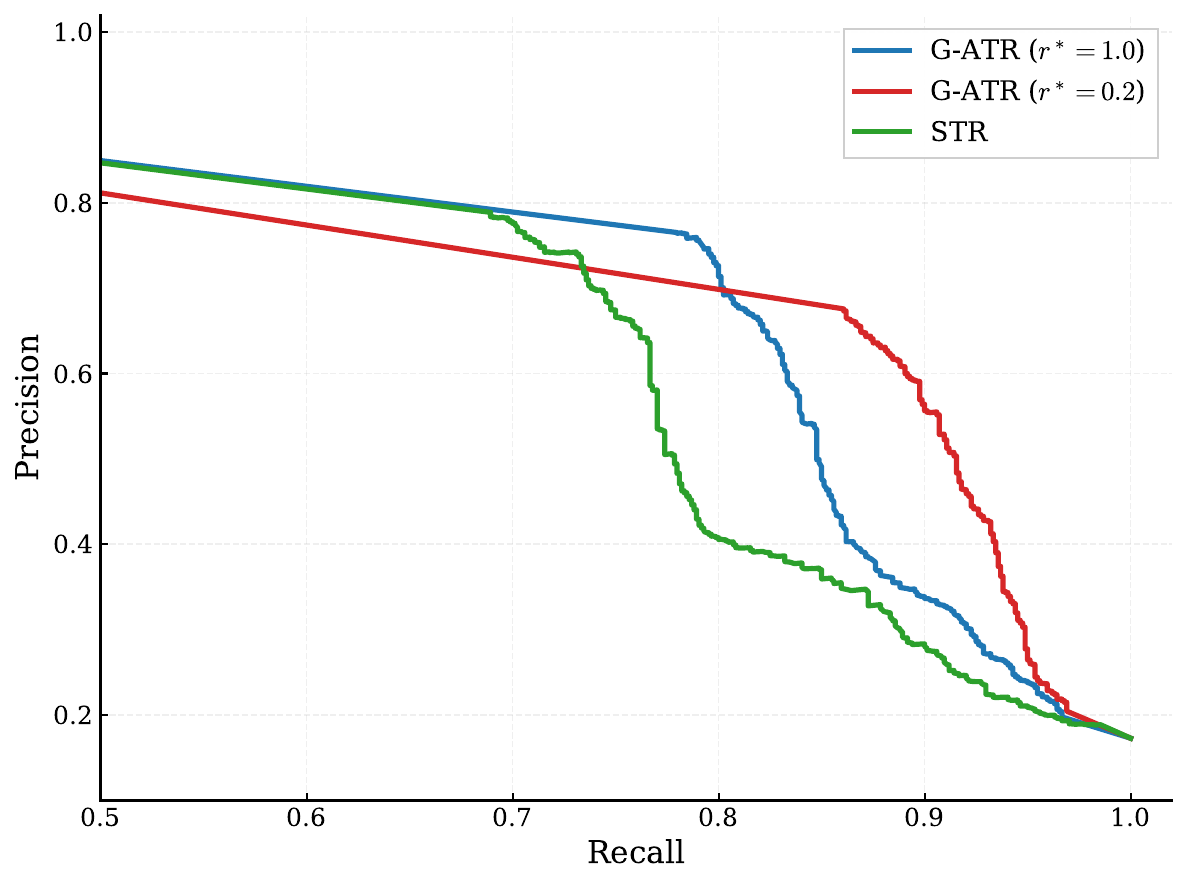}
      \caption{Micro precision-recall curves comparing ATPO (G-ATR) with different target ratios $r^*$ and GRPO (STR). G-ATR with $r^* = 1.0$ improves precision across most recall levels compared with the GRPO model, while a smaller target ratio $r^* = 0.2$ shifts the curve toward higher recall.}
      \label{fig:pr}
  \end{figure}

\section{Precision-Recall Curves}
\label{app:pr_curve}
In Section~\ref{subsec:gcontrol_exp}, we investigate global precision–recall control using the target ratio defined in ATPO. In this section, we provide the corresponding micro precision–recall curves for models trained with the GRPO (STR) and the proposed ATPO (G-ATR) under different target ratios in Figure~\ref{fig:pr}. Compared with the Jaccard reward (green curve), G-ATR with target ratio $r^* = 1.0$ (blue curve) consistently achieves higher precision across most recall levels, indicating a stronger overall performance. When the target ratio is reduced to $r^* = 0.2$ (red curve), the curve shifts toward higher recall regions, demonstrating that the adaptive reward can steer the model toward more recall-oriented operating points while maintaining comparable precision.

\begin{figure}[tb]
  \centering
  \includegraphics[height=7.5cm]{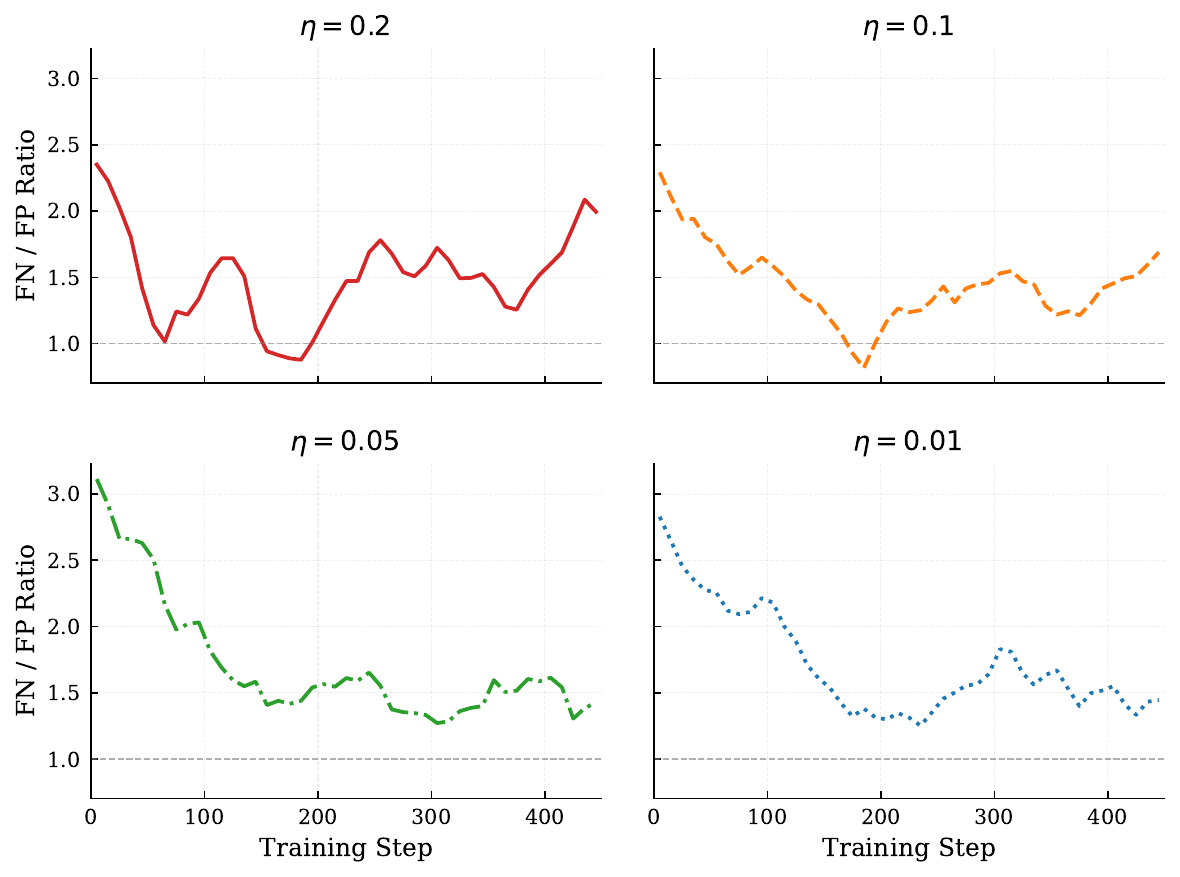}
  \caption{Evolution of Error Type Ratio (ETR=FN/FP) during ATPO training under different step sizes $\eta$, with target ratio $r^* = 1.0$ (dashed line).
  }
  \label{fig:eta}
\end{figure}

\section{Effects of Step Size}
\label{app:hyp_analysis}

We investigate the effects of the step size $\eta$, which determines how strongly the current log-ratio error contributes to the update of $u_{t+1}$ in Equation~\ref{eq:catr_controller}. Figure~\ref{fig:eta} shows the evolution of ETR (FN/FP) on a small validation set as training proceeds under different step sizes $\eta$. The target ratio is fixed at $r^*=1.0$ (dashed horizontal line). Overall, we find that \textbf{ATPO steers the model toward the target ratio.} Across all four settings, ETR starts above the target, indicating a recall-dominant regime. As training progresses, ATPO drives ETR downward toward the target of 1. This suggests that ATPO effectively behaves as a feedback controller, guiding the model toward the specified target.
\\
\textbf{$\eta$ controls controller responsiveness.} For larger $\eta$ (0.2, 0.1), ETR decreases rapidly during early training, which indicates faster convergence and stronger reaction to deviations from the target. However, this stronger control introduces noticeable oscillations. $\eta=0.2$ exhibits larger fluctuations after approaching the target compared to $\eta=0.1$. In contrast, for small $\eta$ (0.05, 0.01), ETR decreases more gradually and convergence takes longer. The resulting trajectories are smoother with dampened oscillations, suggesting better stability of the controller. This indicates that $\eta$ determines how aggressively the controller reacts to instantaneous error. These effects of $\eta$ align with the behavior of a step size parameter in learning and feedback control.

\begin{table}[tb]
  \centering
  \small
  \caption{Training comparison on Qwen3-VL-8B. GRPO uses static rewards
  (Exact Match and STR), whereas ATPO-G and ATPO-C use
  adaptive rewards G-ATR and C-ATR, respectively.
  All RL methods start from SFT-1ep.
  Micro, Macro, and Binary denote F1 scores.
  \textbf{Bold} and \underline{underlining} indicate
  the best and second-best results per column.}
  \label{tab:reward_qwen3}

  \begin{tabular}{@{}lcccc@{}}
    \toprule
    \multirow{2}{*}{Training}
      & \multicolumn{4}{c}{SafeWatch-Bench-Real} \\
    \cmidrule(lr){2-5}
      & Jaccard & Micro & Macro & Binary \\
    \midrule

    Zero-shot
      & 48.69 & 52.45 & 40.32 & 77.76 \\

    SFT-1ep \textit{(RL initialization)}
      & 64.70 & 72.70 & 71.37 & 82.35 \\

    SFT-3ep \textit{(extended SFT)}
      & 74.14 & \textbf{78.07} & \textbf{77.30} & 93.33 \\

    \midrule
    \multicolumn{5}{l}{
      \textit{RL from the same SFT-1ep checkpoint}
    } \\
    \addlinespace[2pt]

    GRPO (EM)
      & 59.95 & 62.50 & 56.79 & 84.33 \\

    GRPO (STR)
      & 67.32 & 68.42 & 65.61 & 86.05 \\

    ATPO-G
      & \textbf{75.50} & 77.08 & 75.25 & \textbf{95.22} \\

    ATPO-C
      & \underline{74.63} & \underline{77.19}
      & \underline{76.09} & \underline{93.79} \\

    \bottomrule
  \end{tabular}
\end{table}

\section{Comparison of Training Objectives on Qwen3-VL-8B}
\label{app:reward_qwen3}

We extend the training objective comparison in Section~\ref{subsec:reward_comp} to Qwen3-VL-8B with results on SafeWatch-Bench-Real reported in Table~\ref{tab:reward_qwen3}. As in the Qwen2.5-VL-7B comparison, all RL methods share the same one-epoch SFT initialization, training data, and GRPO configuration. Consistent with the Qwen2.5-VL-7B results, both ATPO variants outperform the static-reward baselines and their shared initialization across all four metrics. Static-reward RL, however, does not consistently improve over its initialization on this backbone. Compared with extended three-epoch SFT, both ATPO variants achieve higher Jaccard and Binary F1 but slightly lower Micro and Macro F1. Nevertheless, our Qwen2.5-VL-7B experiments show that stronger in-domain performance from extended SFT does not necessarily generalize: extending SFT reduces Jaccard on both SafeWatch-Bench-GenAI (Table~\ref{tab:reward}) and VHD11K-Video-Real~\cite{yeh2024t2vs} (Appendix~\ref{app:ood}).

\begin{table}[tb]
  \centering
  \small
  \caption{Cross-dataset evaluation of Qwen2.5-VL-7B checkpoints
on VHD11K-Video-Real, without
target-dataset fine-tuning. Micro, Macro, and Binary denote
F1 scores. \textbf{Bold} and \underline{underlining} indicate
the best and second-best results per column.}
  \label{tab:ood}

  \begin{tabular}{@{}lcccc@{}}
    \toprule
    \multirow{2}{*}{Model / training}
      & \multicolumn{4}{c}{VHD11K-Video-Real} \\
    \cmidrule(lr){2-5}
      & Jaccard & Micro & Macro & Binary \\
    \midrule



    Zero-shot
      & 54.20 & 38.49 & 34.74 & 63.60 \\

    SFT-1ep \textit{(RL initialization)}
      & \underline{63.64} & \underline{63.12}
      & \underline{50.95} & \underline{80.40} \\

    SFT-3ep \textit{(extended SFT)}
      & 57.63 & 59.29 & 49.08 & 76.00 \\

    \midrule
    \multicolumn{5}{l}{
      \textit{RL from the same SFT-1ep checkpoint}
    } \\
    \addlinespace[2pt]

    GRPO (STR)
      & 59.03 & 58.28 & 42.17 & 78.20 \\

    ATPO-C
      & \textbf{68.50} & \textbf{66.59}
      & \textbf{56.16} & \textbf{84.00} \\

    \bottomrule
  \end{tabular}
\end{table}

\section{Distribution-Shift Evaluation}
\label{app:ood}

Complementing the SafeWatch-Bench-GenAI evaluation in
Table~\ref{tab:reward}, we assess cross-dataset generalization
on the 500 real-world videos from VHD11K~\cite{yeh2024t2vs},
denoted VHD11K-Video-Real. We exclude the generated-video
subset because lower generation quality can make unsafe
content ambiguous. Table~\ref{tab:ood} compares the zero-shot Qwen2.5-VL-7B
backbone with its SafeWatch-Bench-trained SFT and RL
checkpoints, without further fine-tuning on VHD11K.
Both RL methods start from the same SFT-1ep checkpoint.


\noindent\textbf{ATPO improves cross-dataset generalization.} Despite improving in-domain performance (Table~\ref{tab:reward}), extended SFT reduces all four metrics on VHD11K-Video-Real, with Jaccard falling from $63.64$ to $57.63$. GRPO with STR also underperforms SFT-1ep across all four metrics. In contrast, ATPO-C achieves the highest scores across all four metrics, with Jaccard exceeding SFT-1ep and GRPO-STR by $4.86$ and $9.47$ points, respectively. These results support the benefit of adaptive reward optimization for transfer beyond the training distribution.

\begin{table}[tb]
  \centering
  \small
  \caption{Multi-label-only evaluation on SafeWatch-Bench-Real, restricted to videos with at least two ground-truth harmful labels. The SFT, GRPO, and ATPO checkpoints are based on Qwen2.5-VL-7B.}
  \label{tab:multilabel}

  \begin{tabular}{@{}lcccc@{}}
    \toprule
    Model / training & Jaccard & Micro & Macro & Binary \\
    \midrule

    SafeWatch-8B
      & 56.43 & 69.72 & \underline{60.67} & 92.90 \\

    SFT-1ep
      & 28.66 & 43.93 & 31.94 & 76.12 \\

    GRPO (STR)
      & \underline{63.10} & \underline{73.95}
      & 52.19 & \underline{99.39} \\

    ATPO-G-7B
      & \textbf{66.11} & \textbf{76.97}
      & \textbf{66.37} & \textbf{99.70} \\

    \bottomrule
  \end{tabular}
\end{table}

\section{Evaluation on Multi-Label Examples}
\label{app:multilabel}
To examine whether ATPO's gains extend beyond single-label cases, we restrict evaluation to SafeWatch-Bench-Real videos with at least two ground-truth harmful labels. We compare SafeWatch-8B with the Qwen2.5-VL-7B-based SFT-1ep, GRPO (STR), and ATPO-G checkpoints. 

As shown in Table~\ref{tab:multilabel}, ATPO-G-7B achieves the highest scores across all four metrics, improving Jaccard and Macro F1 over GRPO (STR) by $3.01$ and $14.18$ points, respectively. These results show that ATPO's improvements persist on examples with multiple harmful categories and are not confined to single-label cases.

\begin{figure}[!htbp]                                                     
      \centering
      \begin{subfigure}{0.8\textwidth}
          \centering
          \includegraphics[width=\linewidth]{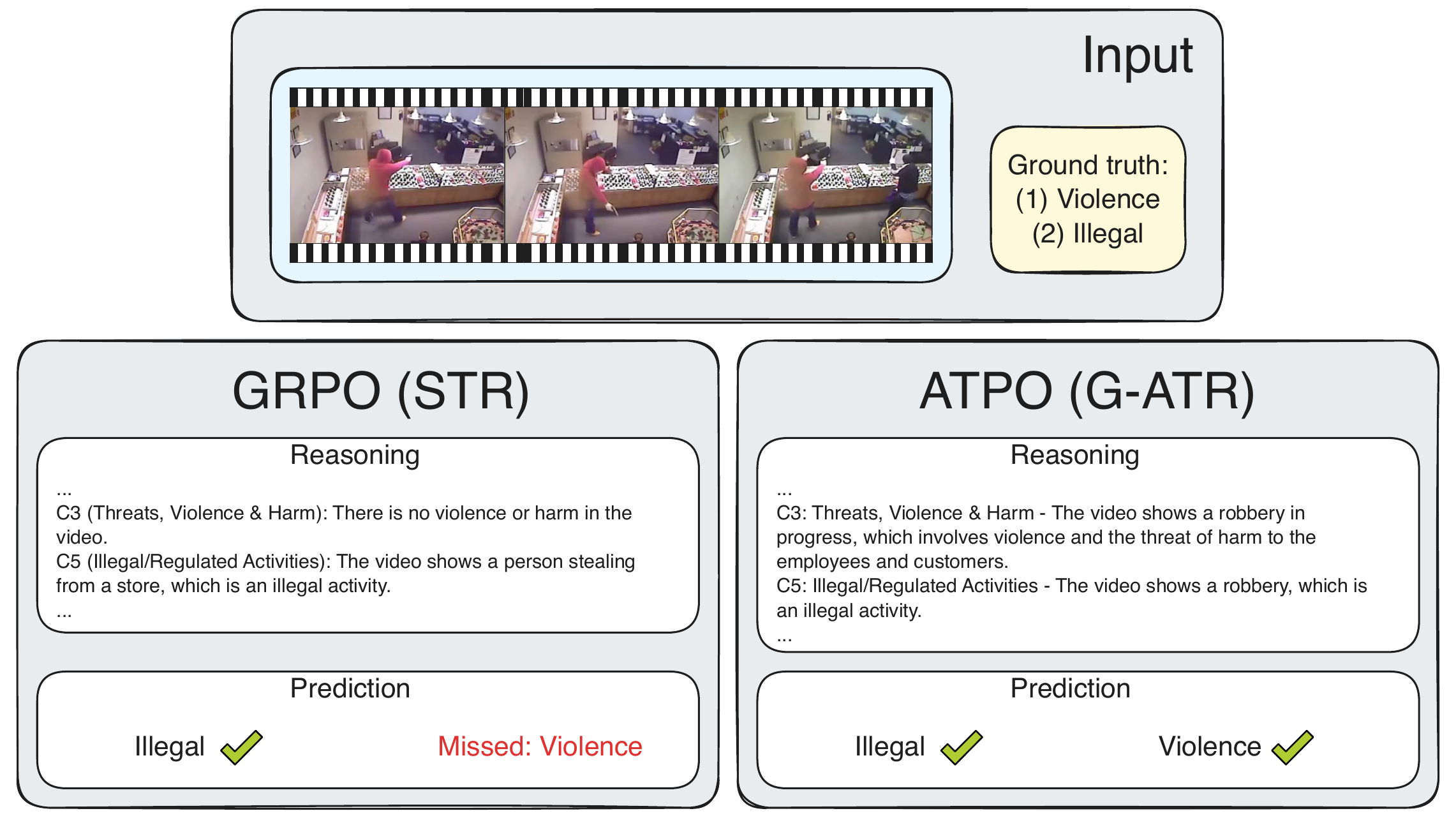}
          \caption{Case 1: a robbery scene.}
          \label{fig:case1}
      \end{subfigure}
      \par\smallskip
      \begin{subfigure}{0.8\textwidth}
        \centering
        \includegraphics[width=\linewidth]{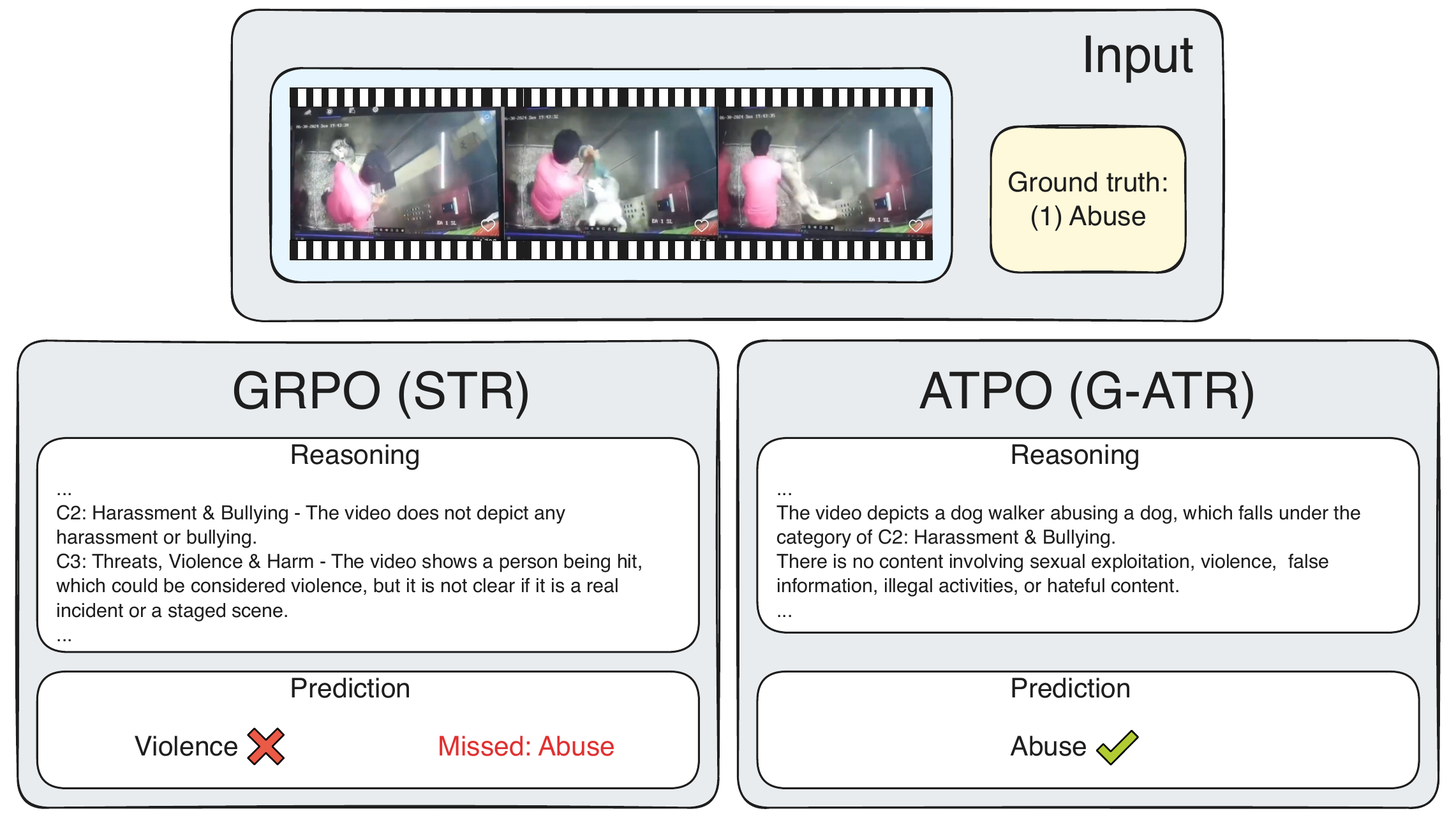}
        \caption{Case 2: a dog walker bullies his dog.}
        \label{fig:case2}
    \end{subfigure}
    \par\smallskip
      \begin{subfigure}{0.8\textwidth}
          \centering
          \includegraphics[width=\linewidth]{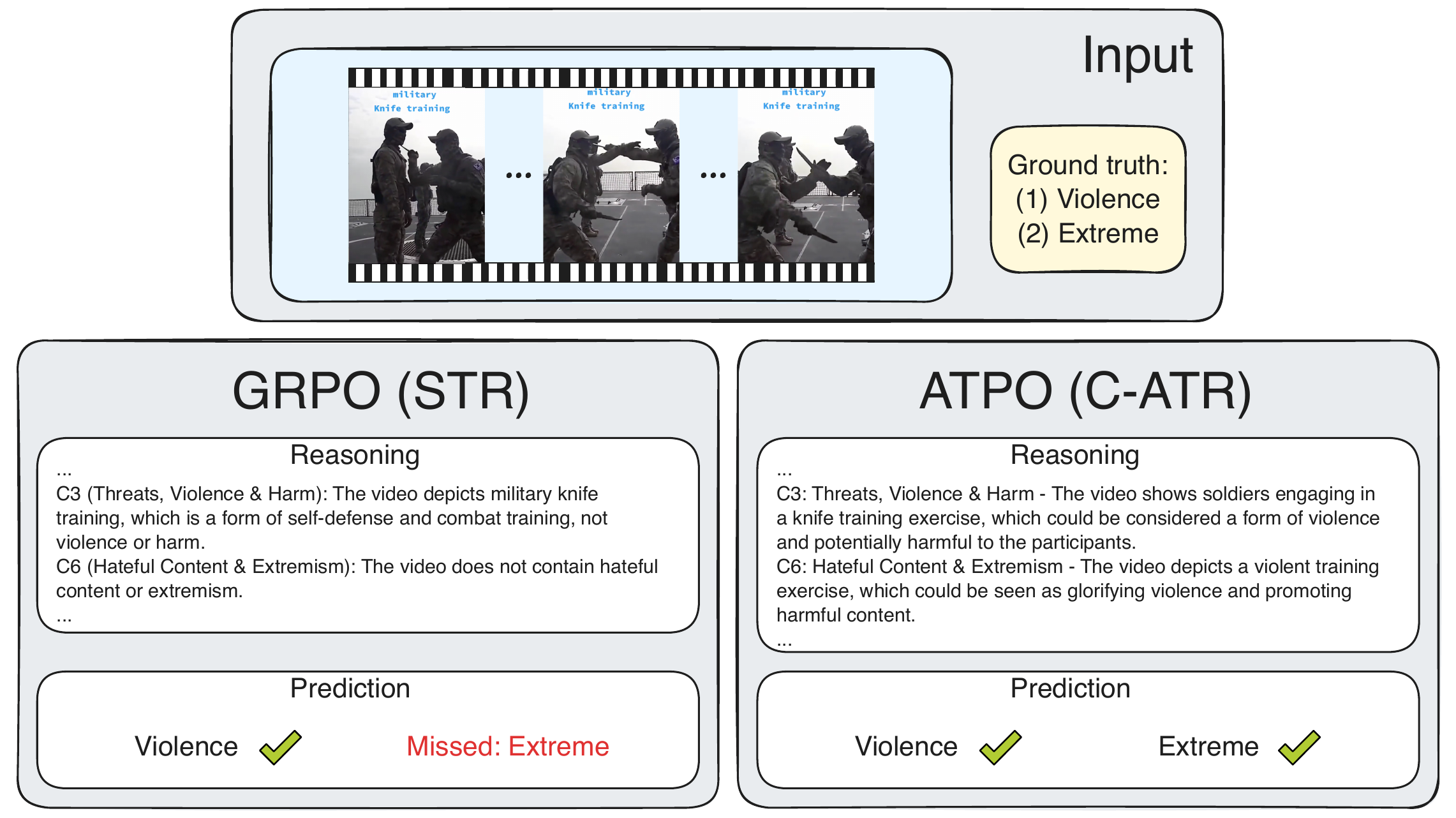}
          \caption{Case 3: a military knife-training video.}
          \label{fig:case3}
      \end{subfigure}
      \caption{Qualitative comparison between GRPO with STR and ATPO with Adaptive Tversky Rewards (G-ATR and C-ATR).}
      \label{fig:cases}
  \end{figure}



\section{Qualitative Analysis}
\label{app:qual}
In this section, we provide qualitative comparisons between models trained with GRPO (STR) and our ATPO framework in Figure~\ref{fig:cases}. Across these examples, the GRPO (STR) model tends to focus on a single salient aspect of the scene, which can lead to missed categories or incorrect interpretations, while ATPO-trained models produce more comprehensive reasoning and more accurate predictions.
\\
In the first case, the video shows a robbery taking place inside a shop. The individuals are attempting to break the display glass while pointing a gun at an employee. The GRPO (STR) model focuses on the theft aspect and predicts only the \textit{Illegal} category while overlooking the violence (shooting) involved. ATPO (G-ATR), on the other hand, recognizes both the illegal activity and the violent threat, resulting in correct predictions.
\\
In the second case, the video shows a dog walker abusing a dog in an elevator, which is labeled as \textit{Abuse}. The GRPO (STR) model produces uncertain reasoning and incorrectly predicts \textit{Violence}, missing the correct \textit{Abuse} category. In contrast, ATPO (G-ATR) correctly identifies the behavior as abuse toward the animal and predicts the appropriate category.
\\
In the third case, the video depicts a military knife training scenario labeled as \textit{Violence} and \textit{Extreme} (with a subcategory of \textit{incitement to violence}). The GRPO model interprets the scene as routine training and therefore dismisses the possibility that the content may reflect or promote harmful or extremist activity. In contrast, the ATPO (C-ATR) model produces reasoning that acknowledges the potentially violent nature of the activity and its potential to glorify violence.
\\
These examples suggest that ATPO can encourage more comprehensive reasoning about harmful content within videos, enabling the model to capture multiple unsafe aspects or correctly interpret the nature of the behavior, thereby reducing missed labels in Multi-VSD.

\section{Complete Category-Level Control Results}
\label{app:category_full}

\begin{table}[tb]
  \centering
  \small
  \setlength{\tabcolsep}{4pt}
  \caption{Complete category-level control results on
  SafeWatch-Bench-Real. Target ratios are modified for
  C4 (Misinformation) and C6 (Extreme); the remaining
  categories are included to examine changes beyond the
  targeted pair. $P$ and $R$ denote precision and recall.}
  \label{tab:category_full}

  \begin{tabular}{@{}lcccccccccc@{}}
    \toprule
    \multirow{2}{*}{Model}
      & \multirow{2}{*}{$r_4^*$}
      & \multirow{2}{*}{$r_6^*$}
      & \multicolumn{4}{c}{C1}
      & \multicolumn{4}{c}{C2} \\
    \cmidrule(lr){4-7} \cmidrule(lr){8-11}
      & & & $P$ & $R$ & $P/R$ & ETR
            & $P$ & $R$ & $P/R$ & ETR \\
    \midrule

    InternVL & -- & --
      & 88.46 & 84.66 & 1.04 & 1.39
      & 56.52 & 18.57 & 3.04 & 5.70 \\
    Qwen3-VL & -- & --
      & 92.31 & 77.01 & 1.20 & 3.58
      & 79.25 & 44.68 & 1.77 & 4.73 \\
    ATPO-C & 1.0 & 1.0
      & 84.69 & 94.15 & 0.90 & 0.34
      & 63.30 & 73.40 & 0.86 & 0.63 \\
    ATPO-C & 5.0 & 0.2
      & 87.95 & 77.67 & 1.13 & 2.10
      & 62.50 & 69.15 & 0.90 & 0.74 \\

    \midrule
    \multirow{2}{*}{Model}
      & \multirow{2}{*}{$r_4^*$}
      & \multirow{2}{*}{$r_6^*$}
      & \multicolumn{4}{c}{C3}
      & \multicolumn{4}{c}{C4 (targeted)} \\
    \cmidrule(lr){4-7} \cmidrule(lr){8-11}
      & & & $P$ & $R$ & $P/R$ & ETR
            & $P$ & $R$ & $P/R$ & ETR \\
    \midrule

    InternVL & -- & --
      & 56.56 & 88.46 & 0.64 & 0.17
      & 94.44 & 19.54 & 4.83 & 70.00 \\
    Qwen3-VL & -- & --
      & 55.47 & 69.34 & 0.80 & 0.56
      & 95.83 & 22.55 & 4.25 & 79.00 \\
    ATPO-C & 1.0 & 1.0
      & 72.44 & 76.53 & 0.95 & 0.81
      & 55.28 & 66.67 & 0.83 & 0.62 \\
    ATPO-C & 5.0 & 0.2
      & 63.91 & 79.81 & 0.80 & 0.45
      & 70.41 & 67.65 & 1.04 & 1.14 \\

    \midrule
    \multirow{2}{*}{Model}
      & \multirow{2}{*}{$r_4^*$}
      & \multirow{2}{*}{$r_6^*$}
      & \multicolumn{4}{c}{C5}
      & \multicolumn{4}{c}{C6 (targeted)} \\
    \cmidrule(lr){4-7} \cmidrule(lr){8-11}
      & & & $P$ & $R$ & $P/R$ & ETR
            & $P$ & $R$ & $P/R$ & ETR \\
    \midrule

    InternVL & -- & --
      & 71.28 & 87.01 & 0.82 & 0.37
      & 94.74 & 15.79 & 6.00 & 96.00 \\
    Qwen3-VL & -- & --
      & 80.49 & 57.89 & 1.39 & 3.00
      & 96.43 & 20.93 & 4.61 & 102.00 \\
    ATPO-C & 1.0 & 1.0
      & 88.68 & 82.46 & 1.08 & 1.67
      & 80.31 & 79.07 & 1.02 & 1.08 \\
    ATPO-C & 5.0 & 0.2
      & 84.76 & 78.07 & 1.09 & 1.56
      & 31.77 & 94.57 & 0.36 & 0.03 \\

    \bottomrule
  \end{tabular}
\end{table}

Table~\ref{tab:category_full} extends the main text category-level control results in Table~\ref{tab:cat_control} to all six categories of SafeWatch-Bench-Real. We report the two ATPO-C configurations with $(r_4^*,r_6^*)=(1.0,1.0)$ and $(5.0,0.2)$, alongside the same zero-shot baselines. The modified configuration targets higher precision for C4 (Misinformation) and higher recall for C6 (Extreme).

The targeted categories shift in the intended directions: C4 precision increases from $55.28$ to $70.41$, while C6 recall increases from $79.07$ to $94.57$, accompanied by a decrease in C6 precision from $80.31$ to $31.77$. For the remaining four categories, the target ratios are held fixed, and their achieved precision-recall ratios remain broadly balanced under both configurations. In particular, P/R changes only slightly for C2 ($0.86$ to $0.90$) and C5 ($1.08$ to $1.09$). C1 and C3 show larger fluctuations, from $0.90$ to $1.13$ and from $0.95$ to $0.80$, respectively, but remain relatively close to one. These results indicate that C-ATR can steer the targeted categories while broadly maintaining the precision-recall balance of categories whose targets are unchanged.

\section{Comparison with Post-hoc Threshold Tuning}
\label{app:threshold}

We compare post-hoc threshold tuning of the Qwen2.5-VL-7B-based
SFT-1ep and GRPO (STR) checkpoints with ATPO-G checkpoints
trained using different global target ratios $r^*$.
The STR checkpoint uses $\alpha=\beta=1$.

\noindent\textbf{Confidence extraction and thresholding.}
Given the input $\mathbf{x}=(x,v)$, each baseline first
generates a complete response $y$ using greedy decoding.
Let $t_c$ denote the token position of the Boolean value
for category $c$. We replay $\mathbf{x}$ and the generated
prefix $y_{<t_c}$ to obtain the next-token probabilities
of \texttt{true} and \texttt{false}, and compute
\begin{equation}
    s_c =
    \frac{
      \pi_\theta(\texttt{true}\mid\mathbf{x},y_{<t_c})
    }{
      \pi_\theta(\texttt{true}\mid\mathbf{x},y_{<t_c})
      + \pi_\theta(\texttt{false}\mid\mathbf{x},y_{<t_c})
    }.
    \label{eq:threshold_score}
\end{equation}
The prefix includes the model's original reasoning and
preceding category decisions. Scores are extracted for
all six categories, including those originally predicted
as false. Samples are included only when all six category
keys appear exactly once and all six scores are valid. Missing scores are not imputed.

Using a single threshold $\tau$ shared across categories,
we construct the predicted label set
\[
    P_\tau = \{c : s_c \geq \tau\}.
\]
The implementation checks that $P_{0.5}=P$, where $P$ is
the label set extracted from the original response.
Lower thresholds can add positive labels, while higher
thresholds can remove them. The original prefixes
$y_{<t_c}$ and scores $s_c$ remain fixed throughout the
sweep. Changed predictions are not propagated into
subsequent category scores.

We sweep $\tau\in\{0.1,0.2,\ldots,0.9\}$ and report five
thresholds in Table~\ref{tab:threshold}. Precision and
recall are micro-averaged over video-category pairs.
The sweep characterizes operating points directly on
the test split, without validation-set threshold selection.


\begin{table}[tb]
  \centering
  \small
  \setlength{\tabcolsep}{4pt}
  \caption{Post-hoc threshold tuning of fixed SFT-1ep and
  GRPO (STR) checkpoints (left), compared with ATPO-G
  checkpoints trained using different target ratios $r^*$
  (right), on SafeWatch-Bench-Real.}
  \label{tab:threshold}

  \begin{tabular}{@{}ccccc@{\hspace{1.5em}}ccc@{}}
    \toprule
    \multicolumn{5}{c}{Post-hoc threshold tuning}
      & \multicolumn{3}{c}{Training-time control} \\
    \cmidrule(lr){1-5} \cmidrule(lr){6-8}

    \multirow{2}{*}{$\tau$}
      & \multicolumn{2}{c}{SFT-1ep}
      & \multicolumn{2}{c}{GRPO (STR)}
      & \multirow{2}{*}{$r^*$}
      & \multicolumn{2}{c}{ATPO-G} \\
    \cmidrule(lr){2-3} \cmidrule(lr){4-5}
    \cmidrule(lr){7-8}
      & Precision & Recall
      & Precision & Recall
      & & Precision & Recall \\
    \midrule

    0.1 & 75.76 & 36.36 & 72.26 & 73.58
      & 0.2 & 66.33 & 86.55 \\
    0.3 & 77.31 & 33.45 & 72.35 & 73.58
      & 0.5 & 76.18 & 80.71 \\
    0.5 & 78.82 & 32.48 & 72.35 & 73.58
      & 1.0 & 75.74 & 78.81 \\
    0.7 & 79.06 & 30.67 & 72.35 & 73.58
      & 2.0 & 75.72 & 69.05 \\
    0.9 & 80.27 & 28.61 & 72.43 & 73.58
      & 5.0 & 78.56 & 67.62 \\

    \bottomrule
  \end{tabular}
\end{table}

\noindent\textbf{Comparison of operating points.}
Across the reported thresholds, SFT-1ep remains in a
high-precision, low-recall regime, while GRPO (STR) exhibits
little change. ATPO-G obtains operating points not recovered
by these threshold settings. At nearly matched precision,
ATPO-G with $r^*=5$ achieves $67.62$ recall, compared with
$32.48$ for SFT-1ep at $\tau=0.5$
(precision $78.56$ versus $78.82$).
Furthermore, ATPO-G with $r^*=0.5$ achieves $76.18$ precision
and $80.71$ recall, exceeding all five reported GRPO (STR)
threshold points in both metrics. These results support
the benefit of training-time adaptive reward optimization
beyond the operating-point adjustments obtained by the
reported global-threshold settings.

\section{Cross-Taxonomy Transfer and Low-Resource Adaptation}
\label{app:taxonomy_transfer}

We examine whether safety knowledge learned on SafeWatch-Bench
transfers to XD-Violence, which uses different category definitions.
Using Qwen2.5-VL-7B as the backbone, we first evaluate zero-shot
transfer of the SafeWatch-Bench-trained ATPO-G-7B checkpoint.
We replace the SafeWatch category definitions in the inference
prompt with those of XD-Violence, without updating the model
parameters. We then adapt this checkpoint for one epoch using G-ATR
on a category-stratified random sample of $395$ XD-Violence
training examples ($10\%$). For reference, we include the
XD-Violence-trained ATPO-G-7B from Table~\ref{tab:main},
obtained through a separate SFT-ATPO pipeline using all
$3{,}950$ target training examples.

\begin{table}[tb]
  \centering
  \small
  \setlength{\tabcolsep}{3pt}
  \caption{Cross-taxonomy transfer and low-resource adaptation
  on XD-Violence. Example counts refer only to XD-Violence
  training data. The final row is a separately trained
  full-data reference.}
  \label{tab:taxonomy_transfer}

  \begin{tabular}{@{}llccccc@{}}
    \toprule
    \multirow{2}{*}{Initialization}
      & \multicolumn{2}{c}{XD-Violence training}
      & \multicolumn{4}{c}{XD-Violence test} \\
    \cmidrule(lr){2-3} \cmidrule(lr){4-7}
      & Method & Examples
      & Jaccard & Micro & Macro & Binary \\
    \midrule

    Backbone
      & None & 0 (0\%)
      & 76.54 & 72.29 & 62.78 & 90.21 \\

    SafeWatch ATPO-G-7B
      & None & 0 (0\%)
      & 80.51 & 76.83 & 64.76 & 94.04 \\

    SafeWatch ATPO-G-7B
      & ATPO & 395 (10\%)
      & 84.58 & 81.00 & 68.00 & 95.10 \\

    \midrule

    Backbone
      & SFT + ATPO & 3,950 (100\%)
      & 88.17 & 85.24 & 71.95 & 96.69 \\

    \bottomrule
  \end{tabular}
\end{table}

As shown in Table~\ref{tab:taxonomy_transfer}, zero-shot
transfer improves all four metrics over the backbone,
with Jaccard increasing from $76.54$ to $80.51$.
Adaptation using only $10\%$ of the target training data
further improves all four metrics, reaching $84.58$
Jaccard, within $3.59$ points of the full-data reference.
These results indicate that safety knowledge learned on
SafeWatch-Bench remains useful under the different category
definitions of XD-Violence and supports adaptation with
limited target-dataset supervision.

\section{Selecting Error-Ratio Targets for Deployment}
\label{app:target_selection}

Error-ratio targets specify the desired balance between missed
harmful content and unnecessary intervention, complementing
the definitions of the safety categories. They are
deployment-specific parameters rather than universal constants.
Practitioners may specify desired precision and recall,
minimum-precision or minimum-recall constraints, or an FN/FP
ratio directly. For category $c$, precision $P_c$ and recall
$R_c$, expressed as proportions, determine the achieved
error ratio:
\begin{equation}
    r_c
    = \frac{\mathrm{FN}_c}{\mathrm{FP}_c}
    = \frac{P_c(1-R_c)}{R_c(1-P_c)}.
    \label{eq:deployment_error_ratio}
\end{equation}

For illustration, a child-facing video service that routes
flagged content to human review might require at least
$80\%$ recall for violence while maintaining at least $60\%$
precision. This prioritizes reducing missed harmful content
while limiting unnecessary review. The boundary operating
point $(P_c,R_c)=(0.6,0.8)$ corresponds to an achieved
error ratio of $r_c=0.375$. These numerical requirements
are illustrative rather than industry standards.

The controller target $r_c^*$ need not equal the desired
achieved ratio, because its relationship with the resulting
precision and recall depends on the model and deployment
distribution. Practitioners should therefore train a small
set of candidate target configurations and evaluate the
resulting checkpoints on held-out labeled data representative
of deployment. Selection should check the actual precision
and recall constraints, rather than error-ratio matching
alone. This procedure applies to both the global target
$r^*$ in G-ATR and the category-specific targets $r_c^*$
in C-ATR. The targets guide training; they are not
inference-time adjustments to a fixed checkpoint.

\section{Societal Impacts}
\label{app:impact}
The proposed ATPO framework is designed to improve automated detection of unsafe content in videos, thereby assisting human moderators in identifying potentially harmful material across multiple categories, such as violence, illegal activities, misinformation, and extremist content. The system is intended to support research on video safety and contribute to the development of more reliable moderation tools for online platforms. In particular, its controllable precision-recall behavior may help adapt detection systems to different deployment requirements, where some scenarios require higher recall to reduce harmful exposure, while others require higher precision to avoid unnecessary intervention.

However, ATPO is not intended to replace human judgment in high-stakes moderation decisions. Its outputs should be used as supporting signals within broader moderation pipelines rather than as final decisions. We believe that improving the reliability and controllability of multi-label video safety detection can contribute positively to online safety by helping moderation systems identify unsafe content more effectively and efficiently.


\section*{NeurIPS Paper Checklist}

\begin{enumerate}

\item {\bf Claims}
    \item[] Question: Do the main claims made in the abstract and introduction accurately reflect the paper's contributions and scope?
    \item[] Answer: \answerYes{} 
    \item[] Justification: Please refer to the abstract, introduction, and conclusion.

\item {\bf Limitations}
    \item[] Question: Does the paper discuss the limitations of the work performed by the authors?
    \item[] Answer: \answerYes{} 
    \item[] Justification: Please see Appendix~\ref{app:limit}.

\item {\bf Theory assumptions and proofs}
    \item[] Question: For each theoretical result, does the paper provide the full set of assumptions and a complete (and correct) proof?
    \item[] Answer: \answerNA{} 
    \item[] Justification: The paper does not include formal theoretical results.

    \item {\bf Experimental result reproducibility}
    \item[] Question: Does the paper fully disclose all the information needed to reproduce the main experimental results of the paper to the extent that it affects the main claims and/or conclusions of the paper (regardless of whether the code and data are provided or not)?
    \item[] Answer: \answerYes{} 
    \item[] Justification: We provide a detailed description of our ATPO algorithm in Section~\ref{sec:method} and Appendix~\ref{app:full_alg}, and details of the experimental setup in Section~\ref{subsec:exp_setup} and Appendix~\ref{app:exp_setup}.

\item {\bf Open access to data and code}
    \item[] Question: Does the paper provide open access to the data and code, with sufficient instructions to faithfully reproduce the main experimental results, as described in supplemental material?
    \item[] Answer: \answerYes{} 
    \item[] Justification: We provide details of the environment in Appendix~\ref{app:exp_setup} and will provide code and scripts upon publication.

\item {\bf Experimental setting/details}
    \item[] Question: Does the paper specify all the training and test details (e.g., data splits, hyperparameters, how they were chosen, type of optimizer) necessary to understand the results?
    \item[] Answer: \answerYes{} 
    \item[] Justification: We describe datasets, splits, models, evaluation metrics, training procedure, and key hyperparameters in Section~\ref{subsec:exp_setup} and Appendix~\ref{app:exp_setup}.

\item {\bf Experiment statistical significance}
    \item[] Question: Does the paper report error bars suitably and correctly defined or other appropriate information about the statistical significance of the experiments?
    \item[] Answer: \answerNo{} 
    \item[] Justification: We note that conducting multiple training runs would require substantially greater computational cost. We found that results are consistent across models and datasets (Section~\ref{subsec:exp_main}, \ref{subsec:reward_comp}, and \ref{subsec:gcontrol_exp}).

\item {\bf Experiments compute resources}
    \item[] Question: For each experiment, does the paper provide sufficient information on the computer resources (type of compute workers, memory, time of execution) needed to reproduce the experiments?
    \item[] Answer: \answerYes{} 
    \item[] Justification: We provide the information in Appendix~\ref{appsub:implement}.
    
\item {\bf Code of ethics}
    \item[] Question: Does the research conducted in the paper conform, in every respect, with the NeurIPS Code of Ethics \url{https://neurips.cc/public/EthicsGuidelines}?
    \item[] Answer: \answerYes{} 
    \item[] Justification: Our research has been conducted with strict adherence to the NeurIPS Code of
Ethics.

\item {\bf Broader impacts}
    \item[] Question: Does the paper discuss both potential positive societal impacts and negative societal impacts of the work performed?
    \item[] Answer: \answerYes{} 
    \item[] Justification: Please see Appendix~\ref{app:impact}.
    
\item {\bf Safeguards}
    \item[] Question: Does the paper describe safeguards that have been put in place for responsible release of data or models that have a high risk for misuse (e.g., pre-trained language models, image generators, or scraped datasets)?
    \item[] Answer: \answerNA{} 
    \item[] Justification: The paper does not release new datasets or high-risk models. Experiments are conducted on existing public benchmarks.

\item {\bf Licenses for existing assets}
    \item[] Question: Are the creators or original owners of assets (e.g., code, data, models), used in the paper, properly credited and are the license and terms of use explicitly mentioned and properly respected?
    \item[] Answer: \answerYes{} 
    \item[] Justification: We use publicly available datasets and models and cite their original sources. We follow their respective licenses and terms of use.

\item {\bf New assets}
    \item[] Question: Are new assets introduced in the paper well documented and is the documentation provided alongside the assets?
    \item[] Answer: \answerNA{} 
    \item[] Justification: The paper does not introduce or release new assets.

\item {\bf Crowdsourcing and research with human subjects}
    \item[] Question: For crowdsourcing experiments and research with human subjects, does the paper include the full text of instructions given to participants and screenshots, if applicable, as well as details about compensation (if any)? 
    \item[] Answer: \answerNA{} 
    \item[] Justification: This paper does not involve crowdsourcing nor research with human subjects.

\item {\bf Institutional review board (IRB) approvals or equivalent for research with human subjects}
    \item[] Question: Does the paper describe potential risks incurred by study participants, whether such risks were disclosed to the subjects, and whether Institutional Review Board (IRB) approvals (or an equivalent approval/review based on the requirements of your country or institution) were obtained?
    \item[] Answer: \answerNA{} 
    \item[] Justification: This paper does not involve crowdsourcing nor research with human subjects.

\item {\bf Declaration of LLM usage}
    \item[] Question: Does the paper describe the usage of LLMs if it is an important, original, or non-standard component of the core methods in this research? Note that if the LLM is used only for writing, editing, or formatting purposes and does \emph{not} impact the core methodology, scientific rigor, or originality of the research, declaration is not required.
    \item[] Answer: \answerNA{} 
    \item[] Justification: We did not use LLMs in developing the core method in this research.

\end{enumerate}

\end{document}